\pdfoutput=1
\documentclass[11pt]{article}
\usepackage[letterpaper,margin=1in]{geometry}
\usepackage[T1]{fontenc}
\usepackage{newpxtext}

\usepackage{amsmath}
\usepackage{amsthm}

\usepackage{amssymb}
\usepackage{newpxmath}
\usepackage{amsfonts}
\usepackage{bbm}
\usepackage{bm}
\usepackage{graphicx}
\graphicspath{{img/}}
\usepackage{grffile}
\usepackage{wrapfig,epsfig}
\usepackage[hyphens]{url}
\usepackage{epstopdf}
\usepackage{enumitem}
\usepackage{booktabs}
\usepackage{multirow}
\usepackage{makecell}
\usepackage[format=plain,labelformat=simple,font=small]{caption}
\usepackage{subcaption}

\usepackage{algorithm}
\usepackage{algorithmicx}
\usepackage{algpseudocode}

\AtBeginDocument{%
}

\usepackage{placeins}
\usepackage{needspace}

\usepackage{tikz}
\usetikzlibrary{spy,calc,arrows.meta,positioning}

\usepackage{adjustbox}
\usepackage{tabularx}
\usepackage{siunitx}

\usepackage[dvipsnames]{xcolor}
\usepackage{tcolorbox}
\usepackage{threeparttable}
\usepackage{array}
\usepackage{colortbl}
\definecolor{bestc}{RGB}{178,205,236}
\definecolor{secondc}{RGB}{219,230,247}
\newcommand{\cb}{\cellcolor{bestc}}
\newcommand{\cs}{\cellcolor{secondc}}
\usepackage{pgfplots}
\pgfplotsset{compat=1.18}

\usepackage[numbers,sort&compress]{natbib}
\usepackage[pagebackref,breaklinks,colorlinks,citecolor=blue,linkcolor=red,urlcolor=magenta]{hyperref}

\theoremstyle{plain}

\theoremstyle{definition}

\newcommand{\R}{\mathbb{R}}

\renewcommand{\textsc}[1]{\textnormal{\scshape #1}}

\newcommand{\cA}{\mathcal{A}}

\newcommand{\cK}{\mathcal{K}}

\newcommand{\cN}{\mathcal{N}}

\newcommand{\cS}{\mathcal{S}}

\makeatletter
\newcommand{\blfootnote}[1]{%
  \begingroup
  \renewcommand{\thefootnote}{}%
  \renewcommand{\@makefntext}[1]{\parindent\z@\noindent##1}%
  \footnote{\raggedright #1}%
  \addtocounter{footnote}{-1}%
  \endgroup
}
\makeatother

\newcommand{\papertitle}{CoRA-NAS: Coarse Ranking and Anchor-Residual Refinement
for Neural Architecture Search}
\title{\papertitle}
\author{%
Yifan Yang\textsuperscript{1*} \quad
Zhaoyan Wang\textsuperscript{2*} \quad
Zheng Gao\textsuperscript{1}\\[0.35em]
Xiaoyu Li\textsuperscript{1} \quad
Jiaojiao Jiang\textsuperscript{1}\\[0.7em]
{\small\textsuperscript{1}University of New South Wales}\\[0.2em]
{\small\textsuperscript{2}Korea Advanced Institute of Science and Technology}}
\date{}
\hypersetup{%
  pdftitle={CoRA-NAS: Coarse Ranking and Anchor-Residual Refinement for Neural Architecture Search},
  pdfauthor={Yifan Yang, Zhaoyan Wang, Zheng Gao, Xiaoyu Li, Jiaojiao Jiang},
  pdfsubject={Neural architecture search and performance estimation},
  pdfkeywords={neural architecture search, zero-cost proxies, learning curves, residual refinement}
}

\begin{document}
\maketitle
\blfootnote{\textsuperscript{*}These authors contributed equally to this work.\\
Corresponding author: Yifan Yang
({\hypersetup{urlcolor=black}\href{mailto:evanyifanyang2026@gmail.com}{\texttt{\bfseries evanyifanyang2026@gmail.com}}}).}

\begin{abstract}
Training-free neural architecture search (NAS) ranks candidate networks from a few
forward/backward passes at initialization, avoiding the cost of full training. Yet zero-cost
ranking has saturated: under one protocol, no single proxy robustly beats the trivial
\#Params/FLOPs baseline across both structure-varying and size-varying spaces. We start from the
observation that a trained network's accuracy is driven by three complementary signals. Two
are static and near-zero-cost: its capacity, and how well its structure uses that capacity at
initialization. The third is its learning dynamics, visible only once training begins and
read here at a small ($\approx$1\% of full training) cost. We realize this as \textbf{CoRA-NAS}
(COarse Ranking + Anchor-residual), a single-configuration, coarse-to-fine method that does not
use fully trained architecture-accuracy labels to fit its ranker
(only the architecture encoding is space-specific, as for any NAS method). Stage~1
(\emph{Rank}) forms an equal-weight rank consensus over existing zero-cost proxies spanning
the two static axes, with a \emph{consensus gate} that adapts the bank per space without ever
reading accuracy. Stage~2 (\emph{Refine}) spends $\approx$1\% of a full-training budget on a
stratified anchor set and corrects the static prior with a target-free early-curve residual.
Across four vision spaces, CoRA-Refine has no weak regime on any of them: its per-space
Spearman is $0.946$ (NB201) / $0.715$ (NB101) / $0.786$ (TransNAS-Bench-101) / $0.894$
(NATS-SSS), so its worst space ($\rho=0.715$) is the highest floor of any compared method,
target-aware LIBRA-NAS included, whereas every baseline collapses on at least one space. The refinement also selects strong architectures, reaching
CIFAR-100 best-found accuracy $73.32$, near the ground-truth best of $73.37$. We are explicit about scope: on the pure size space the unsupervised prior does not
beat \#Params, and the low-cost curve residual corrects this prior, yielding one
configuration robust across all four vision spaces.

\end{abstract}

\section{Introduction}
\label{sec:introduction}

Neural architecture search (NAS) automates the design of network topologies, but its
classical bottleneck is performance estimation: ranking candidate architectures by
how well they will train is what makes search expensive, because the faithful estimator,
training each candidate to convergence, is prohibitively costly over spaces with
$10^4$--$10^{18}$ members~\cite{dong2020nb201,ying2019nb101,zela2022nb301}. \emph{Zero-cost}
(training-free) proxies attack this directly: from a single forward/backward pass at
initialization they produce a score meant to correlate with final accuracy
\cite{abdelfattah2021zerocost,mellor2021naswot,tanaka2020synflow,lin2021zennas}. Two
difficulties, however, are now well documented. First, zero-cost ranking has
saturated: the strongest recent proxies cluster within a narrow band on the home
benchmark and improve on one another only marginally
\cite{lee2024aznas,li2023zico,zhang2022gradsign}. Second, and more damaging, every proxy
has a weak regime. A capacity-driven proxy excels on size-varying spaces but falters
on topology-varying ones; a structure-driven proxy does the reverse. The result is a
\emph{cross-over}: neither a single proxy nor a single simple baseline robustly beats the
trivial \#Params/FLOPs baseline across both structure-varying and size-varying spaces
under one protocol \cite{ning2021evaluating,he2024robot,krishnakumar2022nbsz}. (ZiCo~\cite{li2023zico}
reports beating \#Params, but as a single gradient statistic it is not space-adaptive; we
revisit this on equal footing in \autoref{sec:related-work}.) The community has made the
\#Params baseline an explicit bar to clear, and clearing it everywhere with one
method to close the cross-over remains open.

A natural response is to add training signal. Performance predictors and learning-curve
extrapolators cut cost relative to full training and are often more accurate than any
single zero-cost proxy~\cite{white2021powerful,ru2021sotl,adriaensen2023lcpfn,white2021bananas}.
But the accurate ones are typically supervised: they need ground-truth accuracies
to fit, and a fresh fit (or transfer) per search space. One-shot/supernet methods
introduce their own ranking bias and must be retrained per space
\cite{liu2019darts,guo2020spos}. RoBoT~\cite{he2024robot}, the closest combination method,
robustifies a set of weak proxies but does so with Bayesian optimization that
queries ground-truth evaluations. The gap we target is therefore a method that is
simultaneously (i) \emph{label-free} with respect to fully trained architecture-accuracy labels
used to fit the ranker (early validation curves are used);
(ii) \emph{single-configuration}, with one set of hyperparameters across every dataset and
space (only the architecture encoding is necessarily space-specific, as for any NAS
method); and (iii) a strong cross-space ranker that also selects top architectures
cheaply, across structure-varying and size-varying spaces alike. We are not aware of a prior method
combining all three: the accurate refinement methods query ground-truth accuracy
(RoBoT~\cite{he2024robot}) or are supervised per space, target-aware selectors such as
LIBRA-NAS~\cite{libranas2025} read accuracy to weight proxies, and label-free proxies carry
no refinement at all. The gap is structural, and it is felt most on the size space, where our
unsupervised consensus prior does not beat \#Params, motivating a training-signal correction.

\paragraph{Idea.}
Our starting point is that a trained network's accuracy is driven by three complementary
sources of signal: its capacity (a static budget), how well its structure uses
that budget at initialization (also static), and its \emph{learning dynamics} (dynamic,
visible only once training begins). The two static signals are near-zero-cost; the dynamic
one costs a small fraction of training ($\approx$1\%). Their complementarity is not assumed
but visible in the data: capacity and structure are decoupled by construction in
NATS-Bench, where size spaces isolate the former and topology spaces the latter, so each alone fails
in the opposite regime, which is precisely why no static proxy is robust
everywhere~\cite{dong2021natsbench,ning2021evaluating}.
Static proxies, moreover, cannot observe optimization; the early training curve can.
This motivates a \emph{coarse-to-fine} design: judge a network first on capacity and
structure (static, zero-cost), then correct that judgement with its early training
curve (dynamic, $\approx$1\% cost) exactly where the static prior errs. The dynamic axis sees
optimization the static prior cannot, so the curve residual supplies exactly the signal the
prior structurally lacks rather than re-deriving what it already has. This makes the
correction well-posed rather than redundant (\autoref{sec:method}).

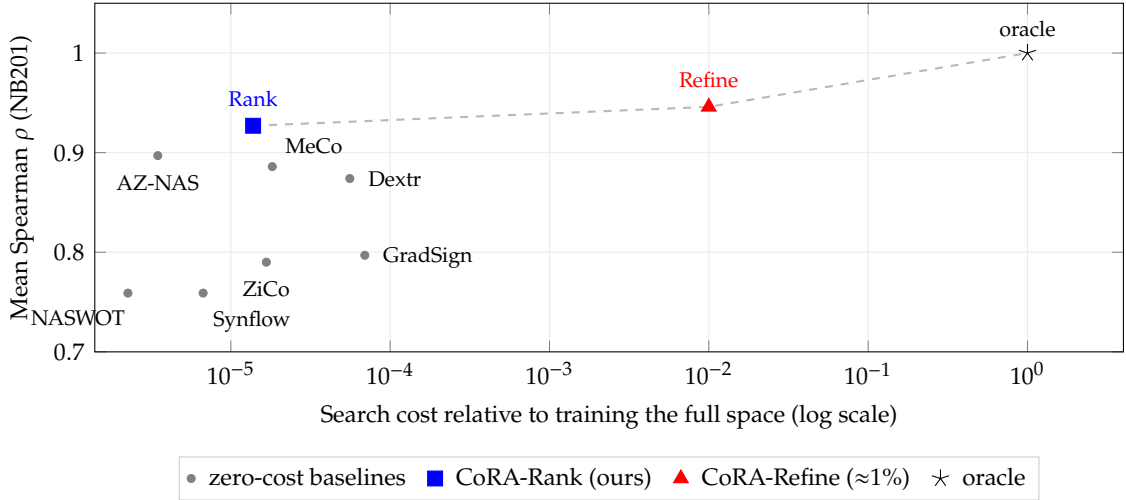
\begin{figure}[t]
\centering
\begin{tikzpicture}
\begin{axis}[
    width=0.92\linewidth, height=6.2cm,
    xmode=log, log basis x=10,
    xlabel={Search cost relative to training the full space (log scale)},
    ylabel={Mean Spearman $\rho$ (NB201)},
    xmin=1.4e-6, xmax=4, ymin=0.70, ymax=1.05,
    xtick={1e-5,1e-4,1e-3,1e-2,1e-1,1},
    ytick={0.7,0.8,0.9,1.0},
    grid=both, grid style={gray!16},
    tick label style={font=\footnotesize},
    label style={font=\footnotesize},
    legend style={font=\footnotesize, at={(0.5,-0.30)}, anchor=north, legend columns=4,
                  draw=gray!40, column sep=0.6ex, /tikz/every even column/.append style={column sep=1.2ex}},
    clip=false,
]
\addplot[dashed, gray!55, line width=0.8pt, forget plot] coordinates {(1.38e-5,0.927) (1e-2,0.946) (1,1.0)};
\addplot[only marks, mark=*, mark size=1.5pt, gray] coordinates {
    (2.26e-6,0.759) (3.48e-6,0.897) (6.70e-6,0.759) (1.67e-5,0.790) (1.82e-5,0.886) (5.58e-5,0.874) (6.93e-5,0.797)
};
\addlegendentry{zero-cost baselines}
\addplot[only marks, mark=square*, mark size=2.8pt, blue] coordinates {(1.38e-5,0.927)};
\addlegendentry{CoRA-Rank (ours)}
\addplot[only marks, mark=triangle*, mark size=3.2pt, red] coordinates {(1e-2,0.946)};
\addlegendentry{CoRA-Refine ($\approx$1\%)}
\addplot[only marks, mark=star, mark size=3.4pt, black] coordinates {(1,1.0)};
\addlegendentry{oracle}
\node[font=\scriptsize, anchor=north] at (axis cs:3.48e-6,0.888) {AZ-NAS};
\node[font=\scriptsize, anchor=north east] at (axis cs:2.5e-6,0.752) {NASWOT};
\node[font=\scriptsize, anchor=north west] at (axis cs:6.70e-6,0.752) {Synflow};
\node[font=\scriptsize, anchor=north] at (axis cs:1.67e-5,0.781) {ZiCo};
\node[font=\scriptsize, anchor=south west] at (axis cs:1.9e-5,0.889) {MeCo};
\node[font=\scriptsize, anchor=west] at (axis cs:6.3e-5,0.874) {Dextr};
\node[font=\scriptsize, anchor=west] at (axis cs:7.8e-5,0.795) {GradSign};
\node[font=\scriptsize, blue, anchor=south] at (axis cs:1.38e-5,0.936) {Rank};
\node[font=\scriptsize, red, anchor=south] at (axis cs:1e-2,0.955) {Refine};
\node[font=\scriptsize, anchor=south] at (axis cs:1,1.006) {oracle};
\end{axis}
\end{tikzpicture}
\caption{\textbf{Cost--quality frontier on NAS-Bench-201.} Mean Spearman $\rho$ (over the
three NB201 datasets, \autoref{tab:nb201}) versus search cost per architecture, measured as
proxy runtime relative to full training (log scale). All zero-cost proxies occupy one
negligible-cost band ($\sim$10--700 ms/arch, ${\sim}10^{-5}$ of full training); within it
CoRA-Rank attains the highest ranking quality ($\rho=0.927$). It is not the cheapest: at
${\approx}140$ ms/arch it runs AZ-NAS's three views plus additional proxies (${\approx}4\times$
AZ-NAS's $35$ ms). But nothing cheaper ranks better, so it sits on the Pareto frontier
(dashed). The $\approx$1\% CoRA-Refine stage then reaches near-oracle $\rho=0.946$; the oracle
is the ground-truth ranking ($\rho=1$).}
\label{fig:teaser}
\end{figure}

\paragraph{CoRA-NAS.}
We instantiate this as a two-stage, label-free method (\autoref{fig:teaser}).
\textbf{Stage~1 (Rank)} percentile-ranks each of a small bank of off-the-shelf zero-cost
proxies covering the two static axes and combines them by an \emph{equal-weight rank
consensus} $\sum_i \log\big(\mathrm{rank}_i(a)/N\big)$, with a \emph{target-free consensus
gate} that adapts the bank per space without reading any label, learning weights, or
tuning on per-space accuracy. \textbf{Stage~2 (Refine)} draws a small stratified anchor set
(a fixed ${\sim}6.4\%$ of the space), trains each anchor for a few epochs, forms a
\emph{target-free} final-accuracy estimate from the early curve, computes the residual
against the prior, and propagates it to all architectures with a tree ensemble before a
light operation-locality smoothing.
The full pipeline uses a single configuration across all datasets and spaces and
introduces no new proxy.

\begin{figure}[t]
\centering
\includegraphics[width=0.86\linewidth]{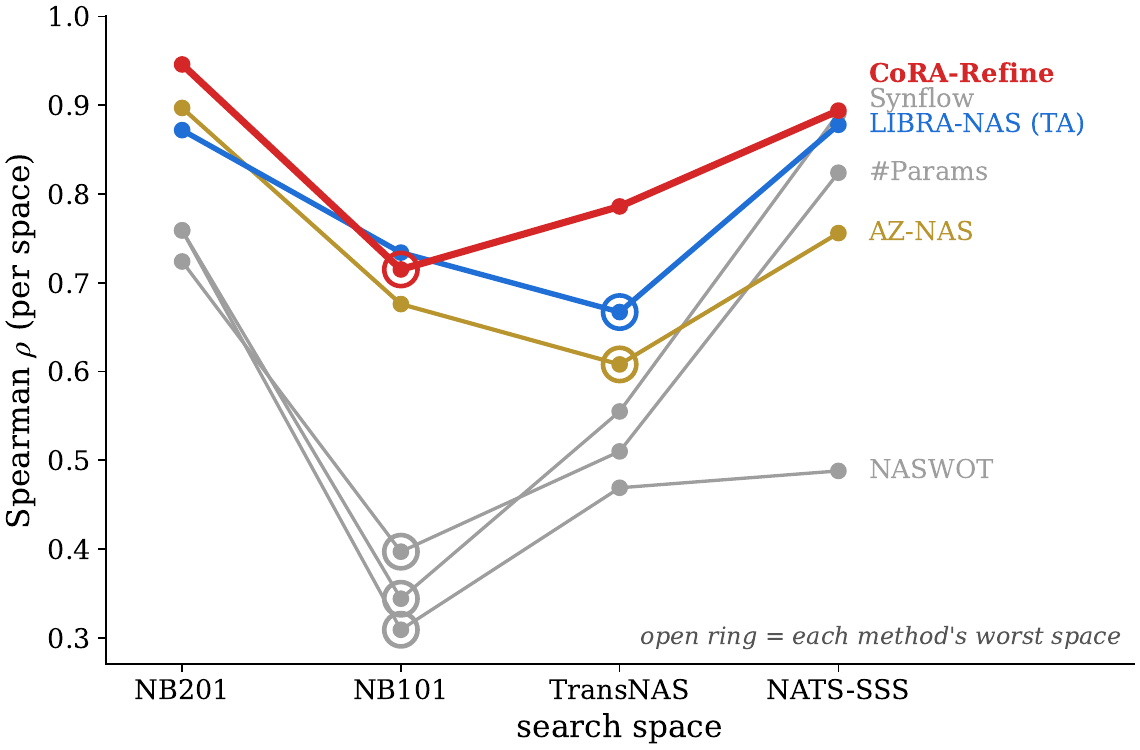}
\caption{\textbf{The cross-over, and how CoRA-Refine closes it.} Per-space Spearman $\rho$
for representative methods across the four vision spaces (values from
\autoref{tab:crossspace}); an open ring marks each method's worst space. Capacity- and
structure-driven proxies each collapse in the opposite regime (grey), and even target-aware
LIBRA-NAS floors at $0.667$ on TransNAS; only CoRA-Refine stays high on all four, with the
highest floor ($\rho=0.715$) of any compared method.}
\label{fig:crossover}
\end{figure}

\paragraph{Findings.}
Our headline is cross-space robustness across four spaces under one configuration
(\autoref{fig:crossover}, \autoref{tab:crossspace}): two topology spaces (NB201, NB101), a
cross-task space (TransNAS-Bench-101), and a size space (NATS-SSS). Together they expose the
cross-over directly: a capacity proxy is strong on size but weak on topology, a structure
proxy the reverse, so every baseline drops sharply on at least one space, and even the
strongest combined proxies floor at $0.55$--$0.61$ somewhere. Across these four vision
benchmarks \textbf{CoRA-Refine has no weak regime}: its worst space ($\rho=0.715$) is the
highest floor of any compared method ($0.946/0.715/0.786/0.894$), target-aware LIBRA-NAS
included. Among label-free methods it has the highest reported means on all four spaces,
with size-space differences from the strongest capacity proxies within noise; the only per-space
value ahead of it anywhere is target-aware LIBRA-NAS on NB101, which
reads accuracy to weight its proxies and still floors lower than CoRA-Refine overall. On the
pure size space the unsupervised consensus prior does not beat \#Params; it is the low-cost,
label-free curve residual that recovers it (\autoref{sec:exp-robustness}). As a pure
zero-cost ranker, CoRA-Rank already leads every baseline on NB201 and on the cross-task
TransNAS space at negligible absolute cost (\autoref{tab:nb201}). NB201 ranking itself is
saturated, so the $\approx$1\% Refine stage adds only a modest ranking margin there by
design; its value is selection (CIFAR-100 best-found $73.32$, near the ground-truth
best of $73.37$ and well above every baseline) and the cross-space recovery (\autoref{sec:exp-crossspace}).

\paragraph{The prior transfers to a real cell space.}
On the real DARTS cell space (via the NB301 surrogate), Refine on real $15$-epoch
curves recovers the capacity-dominated boundary, beating the diluted prior ($+0.046$) and
matching \#Params within noise, with a scramble control confirming the lift is curve-driven.
Trained from scratch ($5$ seeds), its selected cell reaches $2.81\%$ CIFAR-10 error, beating
AZ-NAS and random using the same candidate pool and evaluation protocol ($5$ shared seeds, paired $p<0.015$;
\autoref{sec:exp-search}).

\paragraph{Scope.}
The no-weak-regime result is over four vision spaces, and CoRA is a proxy-agnostic
Rank+Refine \emph{framework} whose benefit scales with the prior's and the encoding's
informativeness. We map its boundary explicitly: on a structurally-distinct recurrent space
(NAS-Bench-NLP) both prior and residual are weak, so CoRA hits an honest weak regime; on a
vision-transformer space (ViT-Bench-101) the label-free consensus is competitive with the ViT
specialist L-SWAG (approximately $0.03$ lower mean Spearman) and ahead of the general ranker AZ-NAS. These
delimit the claim to vision rather than universality (\autoref{sec:exp-crossspace}).

\paragraph{Contributions.}
\begin{itemize}[leftmargin=1.4em,itemsep=2pt,topsep=2pt]
  \item \textbf{C1: Cross-space robustness from one configuration.} A single, untuned
    configuration with no weak regime across four vision benchmarks (NB201, NB101, TransNAS,
    NATS-SSS): the highest worst-case Spearman of any compared method, the highest reported
    means among the non-target-aware methods on the three
    structure/task spaces and, on the size space, matching the strongest capacity proxies while
    decisively beating naive \#Params (\autoref{tab:nb201}, \autoref{tab:crossspace}). A
    \emph{label-free consensus gate} contributes to this robustness, dropping a sign-flipped
    proxy per space from inter-proxy agreement alone, never accuracy, unlike the target-aware
    per-space selection of LIBRA-NAS~\cite{libranas2025} (\autoref{sec:exp-crossspace}).
  \item \textbf{C2: Label-free and budget-tunable.} Fully trained accuracy labels are not
    used to fit the ranker; the only training is a tunable $\approx$1\% anchor budget,
    placing it on an explicit cost--quality frontier (\autoref{fig:teaser}), in contrast to
    label-querying combination methods such as RoBoT~\cite{he2024robot}.
  \item \textbf{C3: A target-free curve-residual refinement.} A two-stage design whose
    second stage corrects the prior's error from early learning curves rather than
    predicting accuracy from scratch, transferable across spaces because it is
    label-free.
  \item \textbf{C4: Clean and interpretable.} An equal-weight consensus over published
    proxies with no learned weights and no new proxy; we show (\autoref{sec:exp-ranking}) that
    learned/adaptive weighting and an opaque off-the-shelf score instead make a
    worse prior. A clean prior is what keeps the curve residual learnable, which
    justifies the simple choice.
\end{itemize}

\section{Related Work}
\label{sec:related-work}

\subsection{Zero-cost (training-free) proxies}
\label{sec:rw-zerocost}
A large family of proxies scores an architecture from a single (or few) forward/backward
pass(es) at initialization. \emph{Gradient-based} proxies include SNIP~\cite{lee2019snip},
GraSP~\cite{wang2020grasp}, Synflow~\cite{tanaka2020synflow},
GradSign~\cite{zhang2022gradsign} and ZiCo~\cite{li2023zico}; \emph{gradient-free}
proxies include NASWOT/jacov~\cite{mellor2021naswot}, the Zen-score~\cite{lin2021zennas}
and the NTK/linear-region view of TE-NAS~\cite{chen2021tenas}; recent entries such as
MeCo~\cite{jiang2023meco}, SWAP~\cite{peng2024swap} and Dextr~\cite{asthana2025dextr}
push correlation higher still on the home benchmark. Two findings from this literature
frame our work. Ning et al.~\cite{ning2021evaluating} show that Synflow is essentially a
smooth parameter counter and that many proxies inherit a strong \#Params bias;
Krishnakumar et al.~\cite{krishnakumar2022nbsz} catalogue 13 proxies over 28 tasks and
report both that the proxies carry \emph{substantial complementary information} and that
none dominates across tasks. ZiCo~\cite{li2023zico} reports being the first single proxy to
consistently beat \#Params; we take that claim seriously and test it on equal footing,
finding that under one unified protocol it is a single gradient statistic that is not
space-adaptive (NB201 mean SPR $0.790$, \autoref{tab:nb201}) and that on a size space the
\#Params cross-over still holds. The recurring conclusion, reinforced by
RoBoT~\cite{he2024robot} and the bias analysis above, is that \textbf{\#Params/FLOPs is
the hard baseline}: no single proxy robustly beats it across both structure and size
spaces under one protocol. We do not propose a new proxy; we reuse published ones, covering
two complementary static axes, combine them without learned weights, then add a cheap
dynamic correction.

\subsection{Aggregating and combining proxies}
\label{sec:rw-aggregation}
Because no single proxy wins, several works combine them. Abdelfattah et al.~\cite{abdelfattah2021zerocost}
``vote'' proxies together; AZ-NAS~\cite{lee2024aznas} assembles four views
(expressivity, progressivity, trainability, complexity) with a non-linear rank
aggregation in a single pass and is the strongest zero-cost ranker we compare against;
RoBoT~\cite{he2024robot} combines weak proxies into a robust metric with Bayesian
optimization and a greedy exploitation step, with theoretical guarantees; LIBRA-NAS
\cite{libranas2025} selects a complementary proxy trio \emph{per space}, but fits that
selection to ground-truth validation accuracy, a target-aware, offline adaptation. The
crucial axis of difference is supervision: RoBoT queries ground-truth evaluations
online and LIBRA-NAS peeks at accuracy offline to choose its proxies, whereas
our consensus and its per-space gate use no labels at all. Relative to AZ-NAS, we use the same off-the-shelf
signals but not its score: we show (\autoref{sec:exp-ranking}) that an
equal-weight rank consensus is not only label-free but makes a \emph{better prior} for
refinement than AZ-NAS's score, which is opaque in the precise sense that its
non-linear aggregation yields a residual the refinement cannot learn from cheap features
(cross-validated $R^2=0.37$ vs.\ $0.43$, \autoref{tab:priorquality}). We do not mean
that the score is undocumented. We also find that learned/adaptive weighting does not help
overall, so we deliberately learn nothing. A second distinction is structural: AZ-NAS, Dextr
and the single proxies above are static metrics with no training-signal refinement
and no mechanism to recover the size space, whereas CoRA adds a label-free curve-residual
second stage that is robust across the cross-over. This unsupervised
consensus-plus-gate-plus-refinement design is what separates us from RoBoT (online labels),
LIBRA-NAS (offline label-fitted per-space proxy selection), AZ-NAS (fixed non-linear score)
and Dextr (single static metric, no refinement).

\subsection{Performance predictors and learning-curve extrapolation}
\label{sec:rw-predictors}
A complementary line predicts final accuracy from partial signal.
White et al.~\cite{white2021powerful} benchmark 31 predictors and conclude that predictors from
different families are composable into stronger estimators, providing methodological support
for our two-stage prior-plus-curve design. The composition they study is of
supervised predictors; our composition is label-free end-to-end (a residual on a
prior, with no ground-truth target), which is what makes it portable across spaces under
one configuration. Learning-curve methods extrapolate a few early
epochs to the final value: classical parametric models (pow$_3$, log-linear/LLW)
\cite{domhan2015lce}, the training-loss surrogate SoTL, which beats validation-accuracy
extrapolation without per-curve tuning~\cite{ru2021sotl}, and LC-PFN, which performs
amortized Bayesian curve extrapolation in a single forward pass~\cite{adriaensen2023lcpfn}.
Supervised predictors such as BANANAS~\cite{white2021bananas} are accurate but require
labelled architectures to train. Our Stage~2 sits in this family but differs in two ways:
it is target-free, forming its per-anchor estimate from the early curve via an
LLW$+$pow$_3$ blend with no ground-truth target, and it predicts a \emph{residual} on the
zero-cost prior rather than accuracy from scratch, which keeps the learnable signal small,
structured and portable across spaces.

\subsection{Multi-fidelity search and the cost--quality frontier}
\label{sec:rw-multifidelity}
Multi-fidelity methods allocate a training budget adaptively: successive halving and
Hyperband~\cite{li2018hyperband}, the Bayesian BOHB~\cite{falkner2018bohb} and
evolutionary DEHB~\cite{awad2021dehb}, and predictor-guided non-uniform halving in
RANK-NOSH, which reaches near-oracle NB201 selection at a fraction of the cost
\cite{wang2021ranknosh}. FEAR~\cite{dey2021fear} freezes training at a fixed error and
notes that zero-cost rankings can degrade as training proceeds. Our Stage~2 occupies the
same cost--quality regime, spending a small, tunable training budget, but is
driven by a label-free curve residual on a fixed anchor set rather than by online
ground-truth comparisons, and Stage~1 provides the zero-cost endpoint of the frontier
(\autoref{fig:teaser}).

\subsection{One-shot NAS and benchmarks}
\label{sec:rw-benchmarks}
One-shot/supernet methods (DARTS~\cite{liu2019darts}, SPOS~\cite{guo2020spos},
ProxylessNAS~\cite{cai2019proxylessnas}, OFA~\cite{cai2020ofa}) amortize training cost but
introduce supernet-induced ranking bias and require retraining per space; we are not a
one-shot method. Evaluation beyond classification encompasses perceptual fidelity and
task-specific reliability, as illustrated by stylization~\cite{GAO2026104895} and
visual-content verification~\cite{gao2026slicesemanticlatentinjection,zeng2026lava}.
Robustness studies of visual-content
verification~\cite{bao2026shiftstochastichiddentrajectorydeflection,10.1145/3774904.3792912}
and graph learning~\cite{wang2025llmsbettergnnhelpers} likewise examine behavior under
changes to inputs or data availability.
For architecture ranking, we evaluate on tabular and surrogate benchmarks that make controlled
comparison possible: NAS-Bench-101~\cite{ying2019nb101} and
NAS-Bench-201~\cite{dong2020nb201} (the latter with full learning curves);
NATS-Bench~\cite{dong2021natsbench}, whose size (SSS) and topology (TSS) variants
decouple capacity from structure and give us a clean cross-space testbed;
TransNAS-Bench-101~\cite{duan2021transnas} for cross-task generalization; and surrogate
learning curves from NAS-Bench-x11~\cite{yan2021nbx11} and the DARTS-space
NAS-Bench-301~\cite{zela2022nb301}.

\section{Method}
\label{sec:method}

\subsection{Problem formulation and a three-axis decomposition}
\label{sec:method-formulation}
Let $\cS$ be a search space of $N=|\cS|$ architectures and let $f(a)$ denote the (unknown,
expensive) final test accuracy of $a\in\cS$. We want a scoring function $\hat s:\cS\to\R$
whose induced ranking matches that of $f$ as closely as possible, so that selecting the
top-ranked architectures recovers near-optimal ones without training the space or
querying any ground-truth accuracy.

Our design rests on the premise that $f(a)$ is driven by three complementary sources of
signal (\autoref{tab:axes}): a static capacity budget, the static quality of an
architecture's structure at initialization (both near-zero-cost), and its
learning dynamics, which are dynamic, visible only once training begins, and read here
at a small ($\approx$1\%) fraction of full-training cost. We argue these axes are complementary, not
redundant, on three grounds.
\emph{(i) Capacity and structure are decoupled by construction.} NATS-Bench isolates them:
its size variant (SSS) fixes topology and varies width, while its topology variant (TSS)
does the reverse~\cite{dong2021natsbench}. Each axis alone therefore fails in the
other regime: capacity dominates size spaces, while structure dominates topology spaces
(\autoref{sec:exp-ranking}). Neither therefore suffices alone.
\emph{(ii) Static cannot observe learning.} Initialization-time proxies structurally cannot
see optimization; the early training curve catches the ``good at init but trains poorly''
(and the reverse) cases that static proxies miss.
\emph{(iii) Information-theoretic.} Zero-cost proxies have been shown to carry substantial
complementary information across tasks~\cite{krishnakumar2022nbsz,ning2021evaluating}.

\begin{table}[t]
\centering
\caption{The three complementary axes of signal that drive final accuracy, and the cheap
estimator CoRA uses for each. The prior (Stage~1) aggregates the two \emph{static} axes;
the refinement (Stage~2) injects the \emph{dynamic} axis as a residual.}
\label{tab:axes}
\small
\setlength{\tabcolsep}{5pt}
\begin{tabular}{@{}l l >{\raggedright\arraybackslash}p{0.40\linewidth} >{\raggedright\arraybackslash}p{0.24\linewidth}@{}}
\toprule
Axis & Nature & What it captures & Cheap estimator \\
\midrule
Capacity          & static  & representational budget (necessary, not sufficient) & \#Params, Synflow \\
Structure         & static  & how the wiring uses that budget at init             & expressivity, trainability, jacov \\
Learning dynamics & dynamic & how it actually optimizes on the data               & early training curve \\
\bottomrule
\end{tabular}
\end{table}

\begin{figure}[t]
\centering
\includegraphics[width=\linewidth]{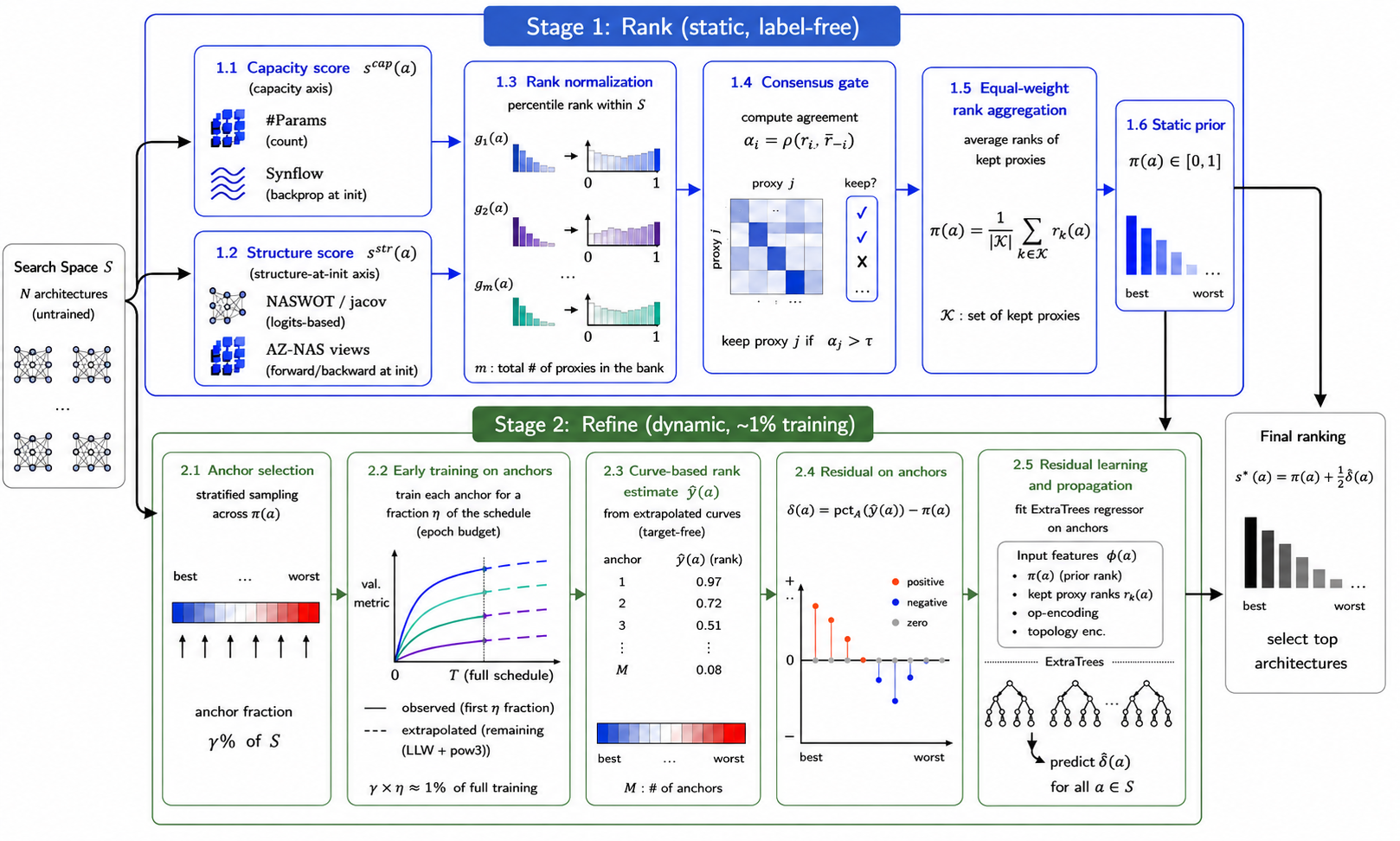}
\caption{\textbf{CoRA-NAS in one pass.} Final accuracy is driven by three cheap, complementary
signals: two \emph{static} axes, \emph{capacity} (\#Params, Synflow) and \emph{structure-at-init}
(NASWOT/jacov, AZ-NAS views), and one \emph{dynamic} axis, \emph{learning dynamics}.
\textbf{Stage~1 (Rank, blue)} percentile-ranks a bank of $m$ off-the-shelf zero-cost proxies,
applies a target-free consensus gate, and aggregates the kept set into a static prior $\pi$.
\textbf{Stage~2 (Refine, green)} draws stratified anchors across $\pi$, trains each for an early
prefix, extrapolates its curve, and propagates an ExtraTrees residual correction.
The diagram is schematic: the implemented prior uses log-percentile aggregation
(\autoref{eq:prior}), and the residual model uses the raw-signal and encoding features in
\autoref{sec:method-refine}, followed by clipping and locality smoothing
(\autoref{eq:combine}--\autoref{eq:smooth}). The anchor and prefix fractions satisfy
$\gamma\eta\approx1\%$ of the cost of fully training the candidate set.
Early validation curves are used; fully trained architecture-accuracy labels are not used to fit the ranker.}
\label{fig:framework}
\end{figure}

\paragraph{Design consequence.}
This decomposition is the spine of a two-stage method. The prior aggregates the two
static axes (Stage~1, \autoref{sec:method-rank}); the refinement injects the
dynamic axis as a residual on that prior (Stage~2,
\autoref{sec:method-refine}). Because the dynamic axis is orthogonal to the static prior,
the curve residual carries exactly the information the prior structurally lacks, yielding a
well-posed and maximally-informative correction, provided the prior is clean enough
that its error remains learnable from cheap features (\autoref{sec:method-justification}).
\autoref{fig:framework} shows the resulting two-stage pipeline end to end.

\subsection{Stage 1: Rank with a zero-cost coarse prior}
\label{sec:method-rank}
Stage~1 produces a label-free coarse ranking from a small bank of $m$ \emph{off-the-shelf}
zero-cost proxies $\{g_i\}_{i=1}^{m}$ that together cover the two static axes: capacity
(\#Params, Synflow~\cite{tanaka2020synflow}) and structure-at-init (the expressivity and
trainability views of AZ-NAS~\cite{lee2024aznas}, and jacov/NASWOT~\cite{mellor2021naswot}).
We introduce no new proxy: the bank is exactly these five signals (\#Params, Synflow,
jacov, and AZ-NAS's expressivity and trainability views). Computing the two AZ-NAS views
reuses AZ-NAS's shared pass, which also yields views we leave unused (e.g.\ progressivity);
the Stage 1 runtime in \autoref{tab:nb201} is the combined cost of all five proxies. All
five are label-free in that they query no ground-truth accuracy, but they are not
\emph{data-free}: jacov reads a small unlabeled real-data minibatch on the
classification spaces (NB201/NB101/NATS-SSS) and a random-normal batch on TransNAS, the two
AZ-NAS views use random-normal inputs, and Synflow uses an all-ones input. CoRA is therefore
label-free/target-free (it never uses accuracy), not data-free; this is the property that
makes it portable across spaces. For each proxy we compute its percentile rank over the space,
\begin{equation}
  r_i(a) \;=\; \frac{\operatorname{rank}_i(a)}{N}\ \in (0,1],
\end{equation}
where $\operatorname{rank}_i(a)$ is the position of $a$ under $g_i$ (best $=N$).

\paragraph{Target-free consensus gate.}
Before combining, we adapt the bank to each space \emph{without any labels} by gating each
proxy on its agreement with the leave-one-out consensus of the others,
\begin{equation}
  \bar r_{-i}(a) = \frac{1}{m-1}\sum_{j\neq i} r_j(a),
  \qquad
  \alpha_i = \rho\big(r_i(\cdot),\,\bar r_{-i}(\cdot)\big),
  \label{eq:gate}
\end{equation}
where $\rho$ is Spearman's correlation taken over $\cS$, and keep the set
$\cK=\{i:\alpha_i\ge\tau_g\}$ with $\tau_g=-0.3$ (a majority floor retains the three
highest-$\alpha_i$ proxies if fewer than three would pass). A proxy that is undefined
on a space (e.g.\ synflow on the attention blocks of a vision transformer, which yields no
finite score) is dropped before the gate is computed, so it cannot corrupt the leave-one-out
agreement; this is a definedness check, not a tuning choice. The gate reads only the
proxies' mutual rank-agreement, never accuracy, and drops a proxy only on a \emph{clear
sign-flip} (it anti-correlates with the rest on this space), not on weak disagreement. The
rationale is a majority-vote argument: when most proxies in the bank carry usable signal on a
space, their consensus is a better ranking than any one of them, and a proxy that
anti-correlates with that consensus is contributing the wrong sign and can only degrade
it. Removing exactly those proxies is therefore expected to improve the consensus without ever
consulting accuracy. This is a label-free heuristic, not a proof that agreement implies accuracy. We
validate it empirically rather than assume it: the gate's only firing (drop jacov on NB101)
coincides with the oracle leave-one-out decision computed against ground truth
(\autoref{sec:exp-crossspace}), and an injected sign-flipped proxy is caught
($\alpha=-0.77$), so the heuristic recovers the label-informed choice on the cases we can check. We
then combine the kept proxies by an equal-weight rank consensus,
\begin{equation}
  s(a) \;=\; \sum_{i\in\cK} \log r_i(a),
  \qquad
  \pi(a) \;=\; \operatorname{pct}_{\cS}\!\big(s(a)\big)\in[0,1],
  \label{eq:prior}
\end{equation}
where higher $s$ is better and $\pi(a)$ is the prior percentile of $a$ within $\cS$.
The $\log$ makes the consensus penalize any proxy ranking $a$ near the bottom (a soft AND);
within $\cK$, equal weighting means \textbf{no learned weights and no per-space tuning on
ground truth}. The gate is a no-op on NB201 and NATS-SSS (all five proxies agree, so
$\cK$ is the full bank and the prior is bit-identical to the ungated equal consensus, leaving
\autoref{tab:nb201} unaffected); on every space we evaluate it keeps the full bank except
NB101, where it removes jacov, which genuinely anti-correlates there
($\alpha_{\text{jacov}}=-0.41$), independently rediscovering jacov as the least transferable
proxy~\cite{abdelfattah2021zerocost} and lifting the NB101 prior (\autoref{sec:exp-crossspace}).
The threshold is robust: any $\tau_g\in[-0.35,-0.2]$ yields the same partition. This is the
key contrast with LIBRA-NAS~\cite{libranas2025}, which selects its per-space proxies by
agreement with \emph{ground-truth accuracy}: our gate adapts per space using only
inter-proxy agreement, so it remains fully label-free. The same gate also filters the raw-signal features of
Stage~2's residual regressor (\autoref{sec:method-refine}), so jacov is dropped
from both the prior and the residual features on NB101, and the bank is unchanged elsewhere. \autoref{sec:exp-ranking} further shows that, on the kept set,
unsupervised adaptive weighting and a 2-axis decoupling both fail to improve the final result
and often hurt it, which justifies equal weighting. Stage~1 trains nothing and
is summarized in \autoref{alg:rank}.

\begin{algorithm}[t]
\caption{Stage 1: Rank (zero-cost coarse prior)}
\label{alg:rank}
\begin{algorithmic}[1]
\Require space $\cS$ ($N=|\cS|$); proxy bank $\{g_i\}_{i=1}^{m}$ over capacity + structure
\For{each $a\in\cS$}
  \For{$i=1,\dots,m$}
    \State $g_i(a)\gets$ one forward/backward pass at init \Comment{no training, no labels}
  \EndFor
\EndFor
\For{$i=1,\dots,m$} \Comment{percentile-rank each proxy over $\cS$}
  \State $r_i(a)\gets \operatorname{rank}_i(a)/N$ for all $a$
  \State $\alpha_i \gets \rho\big(r_i(\cdot),\,\tfrac{1}{m-1}\!\sum_{j\neq i} r_j(\cdot)\big)$ \Comment{leave-one-out agreement; no labels}
\EndFor
\State $\cK\gets\{i:\alpha_i\ge\tau_g\}$ ($\tau_g{=}-0.3$); if $|\cK|<3$, keep the 3 highest-$\alpha_i$ \Comment{consensus gate}
\State $s(a)\gets \sum_{i\in\cK} \log r_i(a)$;\quad $\pi(a)\gets \operatorname{pct}_{\cS}(s(a))$
\Ensure prior percentile $\pi(\cdot)$; kept set $\cK$
\end{algorithmic}
\end{algorithm}

\subsection{Stage 2: Refine with an anchor-based dynamic residual}
\label{sec:method-refine}
Stage~2 spends a tiny training budget to inject the dynamic axis as a correction to the
prior (\autoref{alg:refine}).

\paragraph{Anchors and early curves.}
We draw an anchor set $\cA\subset\cS$, $|\cA|\approx 1000$, stratified across the
prior percentile $\pi$ so that the prior's full range is covered evenly (no clustering).
Each anchor is trained for a few epochs ($5$--$30$, $\approx$1\% of training the space) and
we record its early validation-accuracy curve $c_a=(v_a^{(1)},\dots,v_a^{(T)})$, which
captures the dynamic axis.

\paragraph{Target-free estimate.}
From each curve we form a final-accuracy estimate \emph{without any ground-truth target} by
rank-averaging two parametric extrapolations~\cite{domhan2015lce}: a log-linear (LLW) fit
and a bounded power law (pow$_3$),
\begin{equation}
  \hat y(a) \;=\; \tfrac12\Big(\operatorname{rank}\nolimits_{\cA}\!\big[\mathrm{LLW}(c_a)\big]
  + \operatorname{rank}\nolimits_{\cA}\!\big[\mathrm{pow}_3(c_a)\big]\Big).
\end{equation}
The pow$_3$ term supplies a bounded asymptote that prevents top-end overshoot; both
extrapolations here use the early validation-accuracy curve defined above.

\paragraph{Residual, propagation, smoothing.}
The residual on each anchor is where the prior is wrong,
\begin{equation}
  \delta(a) \;=\; \operatorname{pct}_{\cA}\!\big(\hat y(a)\big) - \pi(a), \qquad a\in\cA .
\end{equation}
We fit an Extremely Randomized Trees regressor~\cite{geurts2006extratrees}
$h:\phi(a)\mapsto\widehat{\delta}(a)$ on the anchors, using features
$\phi(a)=\big[\pi(a),\,\text{Synflow}(a),\,\text{jacov}(a),\,\ell_2\text{-norm}(a),\,\#\text{Params}(a),\,\text{op-encoding}(a),\,\text{topology}(a)\big]$.
These comprise the prior percentile, four cheap raw signals (Synflow log-scaled; the $\ell_2$
weight norm is a capacity cue), and the architecture encoding. A proxy the gate drops
(\autoref{eq:gate}) is likewise removed from these features, so the
sign-flipped jacov is excluded from both the prior and the residual on NB101. We
propagate the residual to every $a\in\cS$, then combine and smooth:
\begin{align}
  \tilde s(a) &= \operatorname{clip}\!\big(\pi(a) + \tfrac12\,\widehat{\delta}(a),\,0,\,1\big),
  \label{eq:combine}\\
  s^\star(a)  &= (1-\beta)\,\tilde s(a) + \beta \!\!\operatorname*{mean}_{a'\in\cN(a)}\!\! \tilde s(a'),
  \qquad \beta=0.3,
  \label{eq:smooth}
\end{align}
where $\cN(a)$ is the set of architectures one operation away from $a$. The half-weight on
the predicted residual hedges against propagation error, and the operation-locality
smoothing in \eqref{eq:smooth} averages each score with its neighbours to denoise the head
of the ranking, exploiting the correlated true accuracy of architectures one edit apart. The final
ranking is by $s^\star$, and we select its top architecture(s). Reported SPR/KT are
computed over the evaluated candidate set $\cS$ (anchors included, ${\approx}6.4\%$ of $\cS$), so the
anchors are not held out of the evaluation; the residual regressor is cross-validated on the
anchors. The two constants here (the residual half-weight and $\beta=0.3$) are held identical across
all datasets and spaces; $\beta$ sits on a wide sensitivity plateau, while the residual
half-weight is the one sensitive axis and is fixed a priori at the cross-space-safe $0.5$
(not the size-space optimum) on a label-free propagation-error argument
(\autoref{sec:exp-robustness}). The
operation-locality smoothing primarily improves head-of-ranking selection (e.g.\ the
CIFAR-10 sample-3000 fix, $93.4\!\to\!94.2$) with negligible effect on whole-space SPR, so
it cannot be inflating the ranking headline.

\begin{algorithm}[t]
\caption{Stage 2: Refine (anchor-based dynamic residual)}
\label{alg:refine}
\begin{algorithmic}[1]
\Require prior $\pi(\cdot)$; kept set $\cK$; features $\phi(\cdot)$ filtered by $\cK$
\Statex \hspace{\algorithmicindent}budget: anchor fraction and epoch-prefix fraction
\State $\cA\gets$ stratified sample of a fixed fraction ($\approx 6.4\%$) of $\cS$ across $\pi$
\For{each $a\in\cA$}
  \State train $a$ for $T$ epochs $\to$ early curve $c_a$ \Comment{the only training; no labels of $f$}
  \State $\hat y(a)\gets \tfrac12\big(\operatorname{rank}[\mathrm{LLW}(c_a)]+\operatorname{rank}[\mathrm{pow}_3(c_a)]\big)$
  \State $\delta(a)\gets \operatorname{pct}_{\cA}(\hat y(a)) - \pi(a)$
\EndFor
\State fit ExtraTrees $h:\phi(a)\mapsto \widehat\delta(a)$ on $\cA$; predict $\widehat\delta(a)$ for all $a\in\cS$
\State $\tilde s(a)\gets \operatorname{clip}(\pi(a)+\tfrac12\widehat\delta(a),0,1)$
\State $s^\star(a)\gets (1-\beta)\tilde s(a)+\beta\operatorname{mean}_{a'\in\cN(a)}\tilde s(a')$
\Ensure refined ranking $s^\star(\cdot)$; select $\arg\max_a s^\star(a)$
\end{algorithmic}
\end{algorithm}

\subsection{Why these choices: a clean prior keeps the residual learnable}
\label{sec:method-justification}
The decomposition of \autoref{sec:method-formulation} only pays off if the residual the
refinement must learn is itself learnable from cheap features. This reframes what makes a
\emph{good prior}: not maximal standalone ranking accuracy, but a prior whose error is
learnable.

\paragraph{Why the residual is learnable from $\phi(a)$.}
The target is not noise but the prior's structured error. By construction
$\delta(a)=\operatorname{pct}_\cA(\hat y(a))-\pi(a)$ is the gap between the dynamic (early-curve)
ranking and the static prior, so it is large exactly where the omitted learning-dynamics axis
diverges from the static rank, and that divergence is not random across the space. Whether a
static proxy over- or under-estimates an architecture's trainability is a property of \emph{which
operations and connectivity} it has (e.g.\ a proxy blind to a normalization or skip pattern
mis-estimates every architecture carrying it), and those are precisely the operation- and
topology-encoding coordinates in $\phi(a)$. The residual is therefore a function of the same
features the regressor sees, which is why it is predictable rather than irreducible. Two existing
controls make this falsifiable, not asserted. \emph{(i) It is real curve signal, not overfitting:}
scrambling which early curve belongs to which anchor collapses the refinement to the prior level
($0.516\!\to\!0.465$ on DARTS, at or below the $0.470$ prior; \autoref{sec:exp-search}). \emph{(ii) Learnability tracks encoding
informativeness, as the mechanism predicts:} where the encoding is only weakly predictive of true
performance, the residual is not learnable and Refine cannot lift the prior. A supervised
encoding$\to$performance diagnostic reads $0.25$ on the recurrent NAS-Bench-NLP space (an honest
weak regime) versus $0.47$ on NATS-SSS, where Refine does recover the size axis
(\autoref{sec:exp-crossspace}). The claim is thus a mechanism with a stated failure mode, not a
guarantee. We make this precise in \autoref{sec:exp-ranking} with an ablation that
feeds different priors into the same refinement (an off-the-shelf opaque score
versus our clean consensus): a clean, signal-transparent equal consensus yields a residual
that the tree ensemble explains substantially better (higher cross-validated $R^2$) and
hence a higher final ranking, whereas an opaque non-linear score, even one that ranks well
on its own, produces an unlearnable residual that the refinement cannot correct.
This is also why we predict a residual rather than accuracy from scratch (the
correction is small and structured) and why the estimate is target-free (it must
run on any space without fully trained performance labels). The budget choices follow the same logic:
a few epochs on a small anchor set, costing $\approx$1\% of full training, suffice because
the limiting factor is predictor quality, not
anchor count, and a head-only variant that trains only the top-ranked anchors makes
selection far cheaper still. Crucially, \textbf{a single configuration}, comprising the proxy bank,
the equal weighting, the gate threshold, the residual weight, $\beta$, and the budget as a
fixed fraction (${\approx}6.4\%$ of architectures trained for ${\approx}15\%$ of the epoch
schedule, i.e.\ ${\approx}6.4\%\times15\%\approx1\%$ of full-training cost), is held fixed across all datasets and spaces in \autoref{sec:experiments}; the
absolute anchor and epoch counts follow from each space's size and training schedule
(e.g.\ $1000$ anchors at $30$ epochs on NB201, $192$ at $14$ on NATS-SSS), and only the
necessarily space-specific architecture encoding (a propagation feature) changes, as
it must for any NAS method.

\section{Experiments}
\label{sec:experiments}

\paragraph{Setup and benchmarks.}
We evaluate on tabular and surrogate NAS benchmarks that permit controlled comparison.
NAS-Bench-201~\cite{dong2020nb201} (15{,}625 architectures, full learning curves) on
CIFAR-10, CIFAR-100 and ImageNet16-120 is our home topology benchmark. For cross-space
generalization we use NATS-Bench~\cite{dong2021natsbench}, whose size variant (NATS-SSS)
decouples width from topology and thus isolates the capacity axis, a second topology space
NAS-Bench-101~\cite{ying2019nb101}, and the cross-task
TransNAS-Bench-101~\cite{duan2021transnas}. CoRA uses a single configuration across
all datasets and spaces (\autoref{sec:method-justification}).

\paragraph{Metrics.}
Following standard practice we report ranking quality as Spearman's $\rho$ (SPR) and
Kendall's $\tau$ (KT) computed over each evaluated candidate set against final test accuracy, and
selection quality as the test accuracy of the selected network under the AZ-NAS
protocol~\cite{lee2024aznas}: best-of-3{,}000-sample selection, mean$\pm$std over $5$ fixed
runs. We lead with SPR/KT (stable, evaluated-set) and report selection accuracy honestly.
Unless a per-dataset value is given, a reported NB201 SPR/KT is the mean over the three
NB201 datasets.

\paragraph{Reproduction protocol (fairness).}
To make the comparison fair we re-run the baselines, except sourced entries noted below,
under one unified protocol on a single NVIDIA L40S, scoring SPR/KT over all 15{,}625 NB201
architectures against test-accuracy at 200 epochs; these re-runs do not copy heterogeneous
published numbers, each computed under its own protocol (some on validation accuracy,
some on a 1{,}000-architecture subsample, some reporting only KT).
Two checks support the port. First, our nine GPU-dependent
proxies match an independent RTX-4090 reproduction within $|\Delta\text{SPR}|\le
0.009$ on all 27 corresponding cells (\#Params and FLOPs are deterministic and
machine-independent; AZ-NAS is not part of the 4090 run), confirming the pipeline is
machine-agnostic. Second, the discrepancies
exceeding $0.02$ from published values are each a documented
protocol or source difference, not a port error (full per-cell table in the appendix). In
particular, T-CET~\cite{tcet} reports KT rather than SPR, so we gate it on KT (our KT matches its paper
to $|\Delta\text{KT}|\le0.018$); and AZ-NAS's published ImageNet16-120 figure is not
reproducible from its released code, so we report our faithfully reproduced value
($\rho=0.877$) and note the discrepancy. Three standard, documented port fixes are applied:
GradSign uses the SUM aggregation, SWAP its regularised SWAP-reg variant, and LIBRA the
RankProduct operator; Fisher and Plain are taken from
NAS-Bench-Suite-Zero~\cite{krishnakumar2022nbsz}.

\subsection{Does one configuration rank well across spaces?}
\label{sec:exp-ranking}

\begin{table*}[t]
\centering
\renewcommand{\arraystretch}{1.18}
\setlength{\tabcolsep}{5pt}
\setlength{\aboverulesep}{0pt}
\setlength{\belowrulesep}{0pt}
\caption{\textbf{Ranking and selection on NAS-Bench-201.} Ranking (KT $\tau$, SPR $\rho$ over all
$15{,}625$ architectures) and selected-network accuracy (mean\,$\pm$\,std, $5$ runs) under one
evaluation protocol; live measurements use a single NVIDIA L40S, while Fisher and Plain
use the precomputed source noted in Appendix~\ref{sec:appendix-additional-material}. Our two stages: \textbf{CoRA-Rank} (Stage~1,
zero-cost prior) and \textbf{CoRA-Refine} (Stage~2, $+{\approx}1\%$ refinement, $^\dagger$).
For KT and SPR, the best value per column is bold with dark-blue shading and the second-best is shaded light blue; proxy input type A/B/F =
architectural/backward/forward. Runtime and budget accounting in \autoref{sec:exp-ranking}.}
\label{tab:nb201}
\resizebox{\textwidth}{!}{%
\begin{tabular}{l c c ccc ccc ccc}
\specialrule{\heavyrulewidth}{0pt}{2pt}
\multirow{2}{*}{Method} & \multirow{2}{*}{Type} & Runtime
 & \multicolumn{3}{c}{CIFAR-10} & \multicolumn{3}{c}{CIFAR-100} & \multicolumn{3}{c}{ImageNet16-120} \\
\cmidrule(lr){4-6}\cmidrule(lr){7-9}\cmidrule(lr){10-12}
 & & (ms/arch) & KT & SPR & Acc. & KT & SPR & Acc. & KT & SPR & Acc. \\
\specialrule{\lightrulewidth}{2pt}{0pt}
\#Params & A & -- & 0.578 & 0.753 & 93.71{\scriptsize\,$\pm$0.17} & 0.552 & 0.728 & 70.73{\scriptsize\,$\pm$0.36} & 0.520 & 0.691 & 42.80{\scriptsize\,$\pm$1.60} \\
FLOPs & A & -- & 0.578 & 0.753 & 93.71{\scriptsize\,$\pm$0.17} & 0.551 & 0.727 & 70.73{\scriptsize\,$\pm$0.36} & 0.519 & 0.691 & 42.80{\scriptsize\,$\pm$1.60} \\
GradNorm & B & 16.4 & 0.355 & 0.483 & 89.05{\scriptsize\,$\pm$0.33} & 0.358 & 0.488 & 61.41{\scriptsize\,$\pm$1.09} & 0.321 & 0.440 & 22.86{\scriptsize\,$\pm$9.96} \\
Grasp & B & 161.4 & 0.312 & 0.450 & 89.10{\scriptsize\,$\pm$0.33} & 0.324 & 0.464 & 62.41{\scriptsize\,$\pm$2.30} & 0.330 & 0.467 & 28.21{\scriptsize\,$\pm$6.40} \\
Snip & B & 140.2 & 0.453 & 0.614 & 89.54{\scriptsize\,$\pm$1.21} & 0.461 & 0.619 & 62.41{\scriptsize\,$\pm$2.30} & 0.403 & 0.539 & 20.81{\scriptsize\,$\pm$9.35} \\
Synflow & B & 67.0 & 0.571 & 0.769 & 93.58{\scriptsize\,$\pm$0.94} & 0.565 & 0.761 & 71.47{\scriptsize\,$\pm$2.52} & 0.555 & 0.747 & 41.95{\scriptsize\,$\pm$5.36} \\
NASWOT & F & 22.6 & 0.556 & 0.742 & 92.05{\scriptsize\,$\pm$1.50} & 0.578 & 0.767 & 68.74{\scriptsize\,$\pm$1.72} & 0.582 & 0.768 & 42.58{\scriptsize\,$\pm$4.03} \\
TE-NAS & B+F & 400.6 & 0.539 & 0.735 & 92.15{\scriptsize\,$\pm$0.16} & 0.526 & 0.717 & 69.92{\scriptsize\,$\pm$1.99} & 0.494 & 0.685 & 44.55{\scriptsize\,$\pm$0.96} \\
ZenNAS & F & 10.7 & 0.296 & 0.386 & 89.59{\scriptsize\,$\pm$1.37} & 0.283 & 0.362 & 64.85{\scriptsize\,$\pm$4.66} & 0.294 & 0.399 & 37.67{\scriptsize\,$\pm$4.05} \\
GradSign & B & 692.5 & 0.622 & 0.812 & 93.71{\scriptsize\,$\pm$0.17} & 0.604 & 0.795 & 70.73{\scriptsize\,$\pm$0.36} & 0.594 & 0.784 & 42.80{\scriptsize\,$\pm$1.60} \\
ZiCo & B & 167.1 & 0.588 & 0.783 & 93.71{\scriptsize\,$\pm$0.17} & 0.603 & 0.797 & 70.73{\scriptsize\,$\pm$0.36} & 0.596 & 0.789 & 42.80{\scriptsize\,$\pm$1.60} \\
AZ-NAS & A+B+F & 34.8 & 0.741 & 0.913 & 93.60{\scriptsize\,$\pm$0.16} & 0.724 & 0.900 & 69.99{\scriptsize\,$\pm$1.04} & 0.695 & 0.877 & 44.90{\scriptsize\,$\pm$1.46} \\
Fisher & B & -- & 0.306 & 0.444 & 91.13{\scriptsize\,$\pm$1.85} & 0.310 & 0.449 & 66.54{\scriptsize\,$\pm$2.47} & 0.315 & 0.457 & 40.60{\scriptsize\,$\pm$3.79} \\
Plain & B & -- & 0.115 & 0.172 & 92.21{\scriptsize\,$\pm$1.82} & 0.106 & 0.158 & 67.28{\scriptsize\,$\pm$3.22} & 0.113 & 0.167 & 41.32{\scriptsize\,$\pm$3.52} \\
MeCo & F & 182.2 & 0.746 & 0.915 & 93.12{\scriptsize\,$\pm$0.95} & 0.720 & 0.895 & 69.42{\scriptsize\,$\pm$1.91} & 0.666 & 0.848 & 42.22{\scriptsize\,$\pm$0.45} \\
SWAP & F & 472.3 & 0.708 & 0.882 & 93.71{\scriptsize\,$\pm$0.17} & 0.696 & 0.873 & 70.73{\scriptsize\,$\pm$0.36} & 0.653 & 0.822 & 42.80{\scriptsize\,$\pm$1.60} \\
T-CET & B+F & 93.5 & 0.644 & 0.824 & 93.99{\scriptsize\,$\pm$0.53} & 0.602 & 0.788 & 72.12{\scriptsize\,$\pm$1.34} & 0.579 & 0.767 & 43.77{\scriptsize\,$\pm$2.38} \\
Dextr & F & 557.6 & 0.720 & 0.898 & 93.31{\scriptsize\,$\pm$0.14} & 0.700 & 0.880 & 70.48{\scriptsize\,$\pm$1.03} & 0.667 & 0.845 & 42.66{\scriptsize\,$\pm$0.70} \\
LIBRA-NAS & A+B+F & 87.8 & 0.698 & 0.876 & 93.56{\scriptsize\,$\pm$0.14} & 0.693 & 0.874 & 70.60{\scriptsize\,$\pm$0.39} & 0.693 & 0.865 & 45.91{\scriptsize\,$\pm$0.43} \\
\midrule
\textbf{CoRA-Rank} (Stage~1) & A+B+F & 138.5 & \cs 0.786 & \cs 0.938 & 93.52{\scriptsize\,$\pm$0.17} & \cs 0.777 & \cs 0.932 & 71.09{\scriptsize\,$\pm$0.63} & \cs 0.760 & \cs 0.911 & 44.24{\scriptsize\,$\pm$0.81} \\
\textbf{CoRA-Refine} (Stage~2) & A+B+F+LC$^\dagger$ & +1k$\times$30ep & \cb\textbf{0.793} & \cb\textbf{0.942} & 94.24{\scriptsize\,$\pm$0.25} & \cb\textbf{0.810} & \cb\textbf{0.951} & 73.32{\scriptsize\,$\pm$0.29} & \cb\textbf{0.802} & \cb\textbf{0.946} & 46.28{\scriptsize\,$\pm$0.24} \\
\midrule
Ground truth & -- & -- & -- & -- & 94.33{\scriptsize\,$\pm$0.05} & -- & -- & 73.37{\scriptsize\,$\pm$0.20} & -- & -- & 47.25{\scriptsize\,$\pm$0.08} \\
\bottomrule
\end{tabular}%
}
\end{table*}

\paragraph{NAS-Bench-201.}
\autoref{tab:nb201} compares CoRA against 19 baselines on the home benchmark. Our zero-cost
Stage~1 (CoRA-Rank) attains the best SPR among all training-free methods on every dataset
(0.938 / 0.932 / 0.911; mean $0.927$) and the best KT (0.786 / 0.777 / 0.760; mean
$0.774$), ahead of the strongest prior zero-cost ranker AZ-NAS (mean SPR $0.897$, KT
$0.720$) and of MeCo~\cite{jiang2023meco} ($0.886$), Dextr~\cite{asthana2025dextr}
($0.874$), LIBRA-NAS~\cite{libranas2025} ($0.872$), SWAP~\cite{peng2024swap} ($0.859$) and
ZiCo~\cite{li2023zico} ($0.790$), all under one protocol (\autoref{sec:experiments}). This
Stage~1 comparison is the like-for-like zero-cost result. Stage~2 (CoRA-Refine) is
not zero-cost: at the $\approx$1\% full-training-cost budget (the $\dagger$ in
\autoref{tab:nb201}) it raises ranking to the top of the table (SPR 0.942 / 0.951 / 0.946;
mean $0.946$), so its bold cells should be read against the low-cost regime, not as a
zero-cost win. On NB201 ranking the field is saturated and this ${\sim}0.02$ SPR margin is
modest by design; Stage~2 earns its budget on selection, raising mean best-found accuracy over CoRA-Rank
on all three datasets and above every baseline, reaching CIFAR-100 best-found $73.32$, close
to the ground-truth best of $73.37$ (mean over $5$ runs; \autoref{tab:nb201}). We do not oversell the single-pick accuracy: on CIFAR-10
the field is compressed near the optimum and several methods (including \#Params-driven
selection) tie within noise; our advantage is clearest in SPR/KT and in the harder
CIFAR-100 / ImageNet16-120 selection.

\begin{table}[t]\centering
\caption{\textbf{Cross-space ranking generalization} (Spearman $\rho$ over the evaluated candidate sets, \emph{one
configuration}). Capacity proxies (\#Params/FLOPs/Synflow) are strong on the size space but weak on
topology; structure proxies (NASWOT) show the opposite; only \textbf{CoRA-Refine} is robust across both.
The first three columns vary structure or task; only NATS-SSS varies capacity. Each cell is a
per-space mean; the per-dataset/per-task breakdown is in Table~\ref{tab:crossspace_detail}
(appendix). Per-cell sourcing,
reproduction caveats, and the target-aware caveat for LIBRA-NAS are detailed in
Appendix~\ref{app:crossspace}.}
\label{tab:crossspace}
\begin{threeparttable}
\setlength{\aboverulesep}{0pt}
\setlength{\belowrulesep}{0pt}
\renewcommand{\arraystretch}{1.15}
\begin{tabularx}{\linewidth}{l *{4}{>{\centering\arraybackslash}X}}
\specialrule{\heavyrulewidth}{0pt}{2pt}
 & \multicolumn{3}{c}{Topology (structure)} & Size \\
\cmidrule(lr){2-4}\cmidrule(lr){5-5}
Method & NB201 & NB101 & TransNAS & NATS-SSS \\
\specialrule{\lightrulewidth}{2pt}{0pt}
\#Params       & 0.724 & 0.397 & 0.510$^{b}$ & 0.824 \\
FLOPs          & 0.724 & 0.397 & 0.522$^{b}$ & 0.512 \\
Synflow        & 0.759 & 0.344 & 0.555$^{b}$ & \cb\textbf{0.891} \\
\midrule
GradNorm       & 0.470 & 0.188 & 0.461$^{b}$ & 0.488 \\
Grasp          & 0.460 & 0.478 & 0.198$^{b}$ & 0.182 \\
Snip           & 0.591 & 0.208 & 0.521$^{b}$ & 0.735 \\
Fisher         & 0.450 & 0.368 & 0.467$^{b}$ & 0.588 \\
NASWOT         & 0.759 & 0.309 & 0.469$^{b}$ & 0.488 \\
ZenNAS         & 0.382 & 0.612 & 0.585$^{b}$ & 0.835 \\
TE-NAS         & 0.712 & 0.296 & 0.295 & 0.205 \\
Plain          & 0.166 & 0.364 & 0.315$^{b}$ & 0.015 \\
\midrule
GradSign       & 0.797 & 0.397 & 0.686 & 0.852 \\
ZiCo           & 0.790 & 0.621 & 0.568 & 0.841 \\
MeCo           & 0.886 & 0.595 & 0.55$^{b,d}$ & \cb\textbf{0.891}$^{n}$ \\
SWAP           & 0.859 & 0.398$^{*}$ & 0.54$^{b}$ & 0.703 \\
T-CET          & 0.793 & 0.397 & 0.638$^{c}$ & 0.770 \\
AZ-NAS         & 0.897 & 0.676 & 0.608 & 0.756 \\
Dextr          & 0.874 & 0.649 & 0.568 & 0.786$^{n}$ \\
LIBRA-NAS$^{TA}$ & 0.872 & 0.734 & 0.667$^{b}$ & 0.878 \\
\midrule
\textbf{CoRA-Rank}              & 0.927 & 0.664 & 0.750 & 0.747 \\
\textbf{CoRA-Refine}$^{\dag}$  & \cb\textbf{0.946} & \cb\textbf{0.715} & \cb\textbf{0.786}$^{a}$ & \cb\textbf{0.894} \\
\bottomrule
\end{tabularx}
\begin{tablenotes}[flushleft]\footnotesize
\item[ ] Bold dark-blue cells mark the best in each column, excluding target-aware LIBRA-NAS; within-noise ties ($\le0.005$) share the marking.
\item[$\dag$] Refine uses a ${\approx}1\%$ prefix-training budget: 1000 anchors for NB201, 192 elsewhere (Appendix~\ref{app:crossspace}).
\item[a] TransNAS anchor curves read from the benchmark's recorded trajectories (${\approx}1\%$ live-equivalent).
\item[b] Sourced from NAS-Bench-Suite-Zero (data-dependent proxies need non-redistributable Taskonomy images).
\item[n] Anti-correlated on NATS-SSS; $|\rho|$ reported (Appendix~\ref{app:crossspace}).
\item[*] SWAP's published NB101 0.77 does not reproduce; we report the reproduced 0.40.
\item[c] T-CET TransNAS is a class\_object surrogate (1 of 3 tasks available). $^{d}$\,MeCo TransNAS 0.55 cited (noise input under-reproduces on dense tasks).
\item[TA] \textbf{Target-aware}: LIBRA-NAS peeks at validation accuracy to pick its per-space trio, so its cells are a soft \emph{upper bound}, not a fair zero-shot baseline.
\end{tablenotes}
\end{threeparttable}
\end{table}

\paragraph{Cross-space generalization.}
\autoref{tab:crossspace} extends the comparison to four spaces under one configuration: two
topology spaces (NB201, NB101), the cross-task TransNAS-Bench-101, and the size space
NATS-SSS. It exposes a cross-over. Capacity proxies (\#Params, Synflow) are strong on the size
space but collapse on topology, while structure proxies (NASWOT) do the reverse and weaken on
size (\autoref{tab:crossspace}). A method leaning on either axis alone therefore fails in the
opposite regime. \textbf{Across these four vision benchmarks CoRA-Refine has no weak regime}:
its worst space ($0.715$, NB101) is the highest worst-case of any method. Every baseline,
including the strong hybrids AZ-NAS and Dextr, drops sharply on at least one space
(\autoref{tab:crossspace}), and CoRA-Refine is the strictly best label-free ranker on the three
structure/task spaces.
(A structurally-distinct non-vision space, NAS-Bench-NLP, is examined below as the honest
boundary of this property.) Even at zero cost, CoRA-Rank
leads all methods on the cross-task space TransNAS and is competitive on NB101
(\mbox{\autoref{tab:crossspace}}), where the target-free consensus gate drops jacov (it
anti-correlates, $\alpha=-0.41$) so the refined score reaches its $0.715$ floor,
the best among fair baselines (AZ-NAS $0.676$). The size space is the honest
boundary: the unsupervised prior is beaten by \#Params (CoRA-Rank $0.747<0.824$), so
it is the cheap, label-free curve residual of Stage~2, not the prior, that recovers it.
There CoRA-Refine ($0.894$) decisively beats naive \#Params ($0.824$; $+0.07$, Holm
$p<0.001$) and ties the strongest capacity proxies Synflow and MeCo ($0.891$;
differences within multiple-comparison-corrected run-noise, \autoref{sec:exp-robustness}).
On the size space the label-free residual thus matches the best size cue rather than
beating it. Since our zero-cost prior does not beat \#Params there, this is the robust
outcome. \autoref{sec:exp-crossspace} decomposes this per space, and
\autoref{tab:crossspace_detail} (appendix) reports the per-dataset/per-task breakdown
behind each per-space mean. Two methodological honesty notes are
folded into the table footnotes. \emph{(i)} LIBRA-NAS is \textbf{target-aware}: it selects
its per-space proxies using ground-truth accuracy, so its cells are a soft upper bound, not
a fair zero-shot baseline; even so, CoRA-Refine exceeds it on three of four spaces (only
NB101, $0.715$ vs.\ $0.734$), and our consensus gate achieves per-space adaptation
without reading any label. \emph{(ii)} Several baselines carry documented per-space
reproduction caveats (SWAP's size-regularised NB101 figure does not reproduce; MeCo's
TransNAS value is cited from its paper; Dextr and MeCo anti-correlate on NATS-SSS, where we
report $|\rho|$); the TransNAS classic baselines are sourced from NAS-Bench-Suite-Zero while
CoRA and the remaining proxies are run live (see the table footnotes and
\autoref{sec:exp-crossspace}).

\paragraph{A clean prior keeps the residual learnable.}
\autoref{tab:priorquality} substantiates the design claim of
\autoref{sec:method-justification} by feeding four different priors into the same
refinement on NB201. The final ranking is driven by two factors at once: a high
starting Rank SPR and a residual the tree ensemble can actually learn. Cross-validated
residual $R^2$ is necessary but not sufficient: it cleanly separates a clean consensus
($R^2=0.43$) from AZ-NAS's opaque non-linear score ($0.37$), but it does not rank
among clean priors: the 3-view consensus has the highest $R^2$ ($0.48$) yet only ties our
final, and params-only has high $R^2$ ($0.46$) yet still loses ($0.862$) because it starts
weak ($0.724$). The decisive contrast is therefore clean-vs-opaque: AZ-NAS is a good
standalone ranker ($0.896$) but a \emph{bad prior}, reaching only $0.929$ through the
identical Refine because its error is unlearnable from cheap features, whereas our clean
consensus starts highest ($0.930$) and ends at the top ($0.951$, tied with the 3-view
consensus). Thus a good prior is judged not by its standalone ranking but by whether it
combines a strong start with a learnable-enough residual, precisely where a transparent
equal consensus wins, and why we learn no weights (C4). Refine is accordingly robust to
which clean prior is used (the 3-view equal consensus reaches the same $0.951$) but
sensitive to clean-vs-opaque, confirming the prior should be kept simple.

\begin{table}[t]
\centering
\caption{\textbf{A clean prior makes a learnable residual.} Four priors fed into the
same Stage 2 refinement (NB201). A good prior combines a strong start with a
learnable-enough residual; CV-$R^2$ separates clean priors ($\ge0.43$) from the opaque
AZ-NAS score ($0.37$) but does not by itself rank among the clean priors (note 3-view's
higher $0.48$ only ties, and params-only's $0.46$ still loses from a weak start). $^{*}$mean
best-found accuracy over the three NB201 datasets. Values are from the exp26 prior-ablation
sweep (run on benchmark learning curves); absolute SPRs differ from the live
\autoref{tab:nb201} by $\le0.006$ (a separate run on cached benchmark curves; the reported
number is the $0.946$ live value), but the prior-vs-prior ordering, the point of this
table, is unaffected.}
\label{tab:priorquality}
\small
\begin{tabular}{l cccc}
\toprule
Prior fed to Refine & Rank SPR & $\to$ Refine SPR & Refine Top-1$^{*}$ & residual CV-$R^2$ \\
\midrule
AZ-NAS score (opaque)        & 0.896 & 0.929 & 70.34 & 0.37 \\
params only (weak)           & 0.724 & 0.862 & 71.39 & 0.46 \\
3-view equal                 & 0.920 & 0.951 & 71.33 & \textbf{0.48} \\
\textbf{CoRA clean consensus}& \textbf{0.930} & \textbf{0.951} & \textbf{71.44} & 0.43 \\
\bottomrule
\end{tabular}
\end{table}

\FloatBarrier
\subsection{Does the refinement select strong architectures, not just rank them?}
\label{sec:exp-search}
\begin{table}[t]\centering
\caption{\textbf{Real-space search on DARTS / CIFAR-10.} All three panels use the DARTS cell
space; margins are small on this low-variance space, and the curve-residual Stage~2 recovers the
capacity-dominated boundary where the zero-cost prior is weakest.
\textbf{(a)} Full-space ranking (Spearman vs.\ the NB301 surrogate) and best-of-10 selection
(oracle $94.42$): the prior alone trails \#Params, and real 15-epoch curves in Stage~2 recover it
(beating diluted Rank $+0.046$, matching \#Params $\Delta{=}{+}0.008$).
\textbf{(b)} Shared-pool selection: all four selectors pick from the \emph{same} $100$k-genotype pool;
the selected networks are trained from scratch under the same protocol at $5$ shared seeds.
CoRA-Refine beats AZ-NAS, CoRA-Rank and random (seed-paired $p<0.015$);
search costs differ as reported below.
\textbf{(c)} Anchor-count ablation: full-space SPR vs.\ stratified-anchor count $K$ on the
$100$k pool ($^\dagger$ nearest the deployed ${\approx}230$). Surrogate SPRs are vs.\ \emph{predicted}
accuracy, not comparable to tabular true-accuracy SPRs (\autoref{sec:exp-search}).}
\label{tab:e5_darts}

{\small\textbf{(a) Surrogate ranking and best-of-10 selection}}\\[2pt]
\begin{tabular*}{0.86\linewidth}{@{\extracolsep{\fill}}l c c}
\toprule
selector & SPR (vs.\ surrogate) & best-of-10 (surr.) \\
\midrule
\textbf{CoRA-Refine} (real 15-ep curves) & \textbf{0.516} & 94.28 \\
\#Params               & 0.508 & 94.30 \\
AZ-NAS                 & 0.477 & 94.26 \\
CoRA-Rank (zero-cost)  & 0.470 & 94.22 \\
Synflow                & 0.193 & 94.08 \\
random search          & --    & 93.82 \\
\bottomrule
\end{tabular*}
\\[6pt]

\label{tab:e15_darts}%
{\small\textbf{(b) Shared-pool selection (trained from scratch, 5 seeds)}}\\[2pt]
{\small
\begin{tabular*}{\linewidth}{@{\extracolsep{\fill}}lcccc}
\toprule
method & search (GPU-d) & surrogate acc & test-err (\%) & params (M) \\
\midrule
random            & $\sim\!0$ & 93.03 & 3.09$\pm$0.10 & 3.23 \\
AZ-NAS            & 0.13 & 93.61 & 3.11$\pm$0.12 & 4.26 \\
CoRA-Rank         & 0.67 & 94.29 & 3.04$\pm$0.06 & 4.14 \\
\textbf{CoRA-Refine} & 1.76 & \textbf{94.55} & \textbf{2.81}$\pm$\textbf{0.05} & 4.56 \\
\midrule
\multicolumn{5}{l}{\emph{reference (gradient / training-free; different paradigm):}} \\
DARTS             & 0.4  & -- & 2.76 & 3.3 \\
PC-DARTS          & 0.1  & -- & 2.57 & 3.6 \\
P-DARTS           & 0.3  & -- & 2.50 & 3.4 \\
TE-NAS            & 0.05 & -- & 2.63 & 3.8 \\
\bottomrule
\end{tabular*}}
\\[6pt]

{\small\textbf{(c) Anchor-count ablation} (CoRA-Rank prior $0.482$, \#Params $0.516$ on the $100$k pool)}\\[2pt]
{\small
\begin{tabular*}{0.86\linewidth}{@{\extracolsep{\fill}}lcccccc}
\toprule
$K$ & 10 & 100 & 200$^\dagger$ & 300 & 600 & 1000 \\
\midrule
SPR & 0.493 & 0.532 & 0.532 & 0.547 & 0.559 & 0.563 \\
\bottomrule
\end{tabular*}}
\end{table}

Stage~2 is a coarse-to-fine search as well as a ranker: because the prior already
concentrates good architectures near its top, we can train only a shortlist of the
highest-ranked anchors and re-rank them by their early curves, a head-only variant
that trades a tiny budget for near-oracle selection. The budget is small and not
tuned: the epoch prefix is a flat plateau from $3$ to $30$ epochs and the anchor count
plateaus above $96$ (\autoref{sec:exp-robustness}), so the ${\approx}1\%$ setting is not a
sweet spot. \autoref{tab:e4_search} makes the cost--quality trade-off concrete on
NATS-SSS/CIFAR-100: ranking the 3{,}000-architecture sample by the zero-cost prior and
re-ranking a top-$N$ shortlist on $K{=}14$-epoch curves, \textbf{CoRA-Rank reaches test
accuracy $70.24$, within $0.5\%$ of the $70.74$ in-shortlist oracle, at ${\approx}2{,}800$
epochs, ${\approx}1\%$ of the $270{,}000$-epoch cost of evaluating the sample}. Against
\emph{prior-free} multi-fidelity search at matched budgets, CoRA-Rank beats
random-search-with-early-stopping and successive halving by $1$--$2\%$ test accuracy at
every budget (\autoref{tab:e4_search}): the uninformed baselines plateau near $68\%$
because more budget merely re-samples the same distribution, whereas a zero-cost prior
focuses the budget on strong architectures. We state two honest points. First, on this
capacity-dominated size space the prior alone is already near-oracle ($70.12$), so the curve
re-rank adds little here ($70.12\!\to\!70.24$). The win is reaching near-oracle
cheaply, not a large re-ranking gain. Second, the search must use a short
best-of-$K$ verify, not the raw Refine argmax: on SSS/CIFAR-100 the Refine score's raw
top-1 over-trusts one misleading early curve and mis-selects ($67.60$) despite its far better
whole-space ranking. This is the SPR-versus-selection gap, which the verified head-only protocol
above avoids. On the home space NB201, where the prior is not near-oracle, the curve
re-rank adds real lift (\autoref{tab:e5_nb201}): across CIFAR-10/100 and $5$ seeds at a
matched epoch budget, at the moderate operating point ($B{=}1{,}200$--$2{,}400$)
CoRA-Rank reaches within ${\sim}0.3$ of the in-shortlist oracle on CIFAR-100
(\autoref{tab:e5_nb201}) and is $\ge$ random search, successive halving, and a synflow-guided
predictor-SHA there; the $12$-epoch re-rank is imperfect on much larger shortlists
(non-monotone at high budget), so we report the moderate-budget point.
\begin{table}[t]\centering
\caption{\textbf{Search efficiency (NATS-SSS / CIFAR-100).} Best-found test-acc@90 vs.\ epoch budget. A
zero-cost prior (CoRA-Rank) reaches within $0.5\%$ of the $70.74$ in-shortlist oracle at $\approx$1\% of the
$270{,}000$-epoch full-evaluation budget, and beats prior-free multi-fidelity search by $1$--$2\%$ at every
matched budget. The NB201 frontier and real DARTS search are reported in
\autoref{tab:e5_nb201} and \autoref{tab:e5_darts}, respectively.}
\label{tab:e4_search}
\setlength{\aboverulesep}{0pt}
\setlength{\belowrulesep}{0pt}
\renewcommand{\arraystretch}{1.15}
\begin{tabularx}{0.82\linewidth}{r *{3}{>{\centering\arraybackslash}X}}
\specialrule{\heavyrulewidth}{0pt}{2pt}
budget (ep) & \textbf{CoRA-Rank} & random$+$early-stop & successive-halving \\
\specialrule{\lightrulewidth}{2pt}{0pt}
 350  & \cb\textbf{69.88} & $68.95\pm1.04$ & 68.43 \\
 700  & \cb\textbf{69.88} & $68.96\pm1.00$ & 67.58 \\
1400  & \cb\textbf{69.88} & $68.15\pm0.76$ & 68.20 \\
2800  & \cb\textbf{70.24} & $68.54\pm1.30$ & 68.23 \\
5600  & \cb\textbf{70.24} & $68.13\pm0.89$ & 68.54 \\
11200 & \cb\textbf{70.24} & $68.07\pm0.81$ & 68.13 \\
\bottomrule
\end{tabularx}
\end{table}

\begin{table}[t]\centering
\caption{\textbf{End-to-end search on the home space (NB201, CIFAR-10/100).} Best-found test accuracy at a
\emph{matched} epoch budget (5 seeds for the stochastic baselines). At the moderate-budget operating point
($B{=}1200$--$2400$) CoRA-Rank reaches within $\sim$0.3 of the in-shortlist oracle and is $\ge$ random search,
successive-halving and a synflow-guided predictor-SHA. The 12-epoch re-rank is imperfect on larger shortlists
(non-monotone at higher budget), so we report the moderate-budget point. (NB201 ImageNet16-120 signals are not
cached; a from-scratch DARTS run is reported in Table~\ref{tab:e5_darts}.)}
\label{tab:e5_nb201}
\begin{tabular*}{\linewidth}{@{\extracolsep{\fill}}l r cc ccc c}
\toprule
dataset & $B$ (ep) & rank-only & \textbf{CoRA-Rank} & random+ES & SHA & pred-SHA & oracle \\
\midrule
CIFAR-10  & 1200 & 93.89 & \textbf{94.04} & 93.05 & 93.30 & 93.99 & 94.10 \\
CIFAR-10  & 2400 & 93.89 & \textbf{94.04} & 93.27 & 93.74 & 93.99 & 94.37 \\
CIFAR-100 & 1200 & 71.18 & 72.42 & 70.38 & 71.51 & 71.60 & 72.42 \\
CIFAR-100 & 2400 & 71.18 & \textbf{73.22} & 71.01 & 71.95 & 71.60 & 73.51 \\
\bottomrule
\end{tabular*}
\end{table}

\paragraph{A real-space search on DARTS.}
Beyond the tabular benchmarks, we run an end-to-end search on the real DARTS cell space
via the NB301 surrogate (\autoref{tab:e5_darts}): $3{,}000$ genotypes scored by each method,
shortlisted, and surrogate-evaluated (best-of-$10$; oracle $94.42$). DARTS is
capacity-dominated, like NATS-SSS, so the zero-cost prior alone (CoRA-Rank, surrogate-Spearman
$0.470$; here and below correlations are against the NB301 surrogate's predictions, a noisier
reference than tabular true accuracy) matches AZ-NAS ($0.477$) but trails \#Params ($0.508$); diagnostically, the
equal-weight consensus is diluted by two uninformative proxies (jacov, trainability;
\autoref{sec:limitations}). CoRA's two-stage method recovers it: feeding \textbf{real 15-epoch
CIFAR-10 curves} from ${\sim}120$ shortlisted cells into Stage 2 significantly beats
the diluted prior (Refine $0.516$ vs.\ Rank $0.470$, $+0.046$) and the strongest zero-cost
ranker AZ-NAS ($0.477$; paired bootstrap, Holm $p{=}0.008$), and ties \#Params (paired
$\Delta{=}{+}0.008$, $p{=}0.60$, 95\% CI $[-0.018,+0.031]$; within noise, the same standard
we apply on NATS-SSS), at $\sim$a few GPU-hours versus DARTS's ${\sim}1.5$ GPU-days to fully
train one cell. A scramble control that shuffles which curve belongs to which anchor
collapses Refine to at or below the prior ($0.516\!\to\!0.465$, versus the $0.470$ prior; within
noise of it), confirming the lift is driven by the real training curves, not feature
overfitting. As elsewhere the gain is in ranking;
best-of-$10$ selection is a near-tie with \#Params at this budget
(\autoref{tab:e5_darts}). We bound this
honestly: the Spearman here is against the surrogate's predictions (a compressed,
noisier reference, not comparable to the tabular true-accuracy SPRs of
\autoref{tab:crossspace}), so we also run a real shared-pool search comparison and train the
selected cell from scratch (\autoref{tab:e15_darts}): scoring a $100$k-genotype pool and
training CoRA-Refine's pick under the standard DARTS evaluation ($600$ epochs, $5$ seeds), it
reaches $\mathbf{2.81\pm0.05\%}$ test error ($97.19\%$), trailing the best gradient/training-free
references ($2.50$--$2.76\%$) but beating the other shared-pool selectors: with all four picks trained at five
shared seeds, CoRA-Refine ($2.81{\pm}0.05\%$) beats AZ-NAS ($3.11{\pm}0.12$), CoRA-Rank
($3.04{\pm}0.06$) and random ($3.09{\pm}0.10$), seed-paired and significant (paired-$t$
$p<0.015$, all five seeds favouring CoRA-Refine). It is not a Pareto win: CoRA-Refine's
$1.76$ GPU-day search is dearer than the gradient/training-free references and does not undercut
their $2.50$--$2.76\%$ (a different search paradigm, shown for context); margins are small on
this low-variance space (CoRA-Rank and random tie), so the surrogate-SPR ranking remains the
more reliable comparator. An anchor-count ablation
confirms the curve component is real: Refine's SPR rises monotonically with the anchor budget,
overtaking \#Params ($0.516$) by $K{\approx}100$ and plateauing (${\sim}0.56$) by $K{\approx}600$
(\autoref{tab:e15_darts}); the deployed cell uses ${\approx}230$ anchors, on the rising part just
past the \#Params-crossing knee, trading a little ranking headroom for lower cost. These are two
distinct experiments: the ${\sim}120$ shortlisted cells above are the best-of-$10$ pool of the
$3{,}000$-genotype \emph{surrogate} run, whereas the ${\approx}230$ anchors (nearest tabulated
grid point $K{=}200^\dagger$) are the Stage-2 budget of the $100$k-pool \emph{shared-pool} comparison. Still, it makes the thesis concrete on a real
space: the dynamic curve residual recovers exactly the capacity-dominated boundary where
the static prior is weakest.

\FloatBarrier
\subsection{Where does each stage close the cross-over?}
\label{sec:exp-crossspace}

We unpack \autoref{tab:crossspace} along two axes: how the gate adapts the prior per
space, and how Refine repairs each space.

\paragraph{The consensus gate adapts the prior without labels.}
The target-free gate of \autoref{eq:gate} measures each proxy's leave-one-out agreement with
the others on the current space and drops only a clear sign-flip. It runs on every
space-instance behind \autoref{tab:crossspace} and fires exactly once: on NB101, jacov
anti-correlates with the consensus ($\alpha_{\text{jacov}}=-0.41$) and is dropped, lifting
the prior from $0.628$ to $0.664$. On all nine other instances, namely NB201 ($\times3$), NATS-SSS
($\times3$, including ImageNet16-120), and the three TransNAS tasks, every proxy clears the
threshold (the lowest, TransNAS expressivity at $-0.17$, is correctly kept, a weak
disagreement that should not trigger a drop), so the bank is unchanged and the prior
is bit-identical to the ungated consensus (NB201 in \autoref{tab:nb201} is unaffected).
Because the gate never reads accuracy, its decision on TransNAS is task-agnostic (identical
across Object/Scene/Jigsaw), and the $-0.41$-versus-$-0.17$ gap means any
$\tau_g\in[-0.35,-0.2]$ gives the same partition, so the gate is not a tuned knob; it
independently rediscovers jacov as the least transferable
proxy~\cite{abdelfattah2021zerocost}, using only inter-proxy agreement and never
accuracy, in contrast to LIBRA-NAS, whose per-space selection is fit to ground-truth labels.
The gate's role is to remove a genuine sign-flip without labels, not to win a space at zero
cost: even after it, zero-cost CoRA-Rank on NB101 ($0.664$) sits just below AZ-NAS ($0.676$),
and the label-free NB101 lead is delivered by Refine ($0.715$). That the single firing is a
detector rather than a hard-coded case is confirmed by a deliberately injected sign-flipped
proxy, which is also caught and dropped ($\alpha=-0.77$) while all five real proxies are
kept. The gate activates whenever a proxy genuinely flips; it simply rarely does.

\paragraph{Refine repairs every space.}
Reading \autoref{tab:crossspace} as Rank$\to$Refine, the residual lifts every space:
NB201 $0.927\!\to\!0.946$, NB101 $0.664\!\to\!0.715$, TransNAS $0.750\!\to\!0.786$, and
most of all the size space NATS-SSS $0.747\!\to\!0.894$, exactly where the static prior is
weakest. This is the core claim made concrete: a single label-free residual makes one
configuration robust across structure, size, and task spaces. Notably, on the cross-task
TransNAS benchmark CoRA's prior uses random-normal inputs throughout and reads no
Taskonomy task images, so this cross-task result needs neither labels nor task data, a
sharper statement than label-free alone.

\paragraph{The cross-space lead is not a provenance artifact.}
\autoref{tab:crossspace}'s TransNAS row mixes sources (classic proxies from
NAS-Bench-Suite-Zero, MeCo cited; see its footnotes). To rule out a mixed-provenance
artifact we recompute every proxy live under one recipe on the same
$3{,}000$-architecture TransNAS/class\_object sample (\autoref{tab:e9_live}): CoRA-Rank
($0.693$) and CoRA-Refine ($0.732$) lead the best live baseline, AZ-NAS ($0.631$), by
$+0.06$/$+0.10$, so the cross-space advantage survives a strict single-provenance comparison.
(These single-task values sit below the Object/Scene/Jigsaw mean in \autoref{tab:crossspace};
the ordering is the point.)
\paragraph{Honest size-space caveat.}
On the pure size space the gain is in ranking, not single-pick selection. The static
prior is already a near-oracle selector there (its top-1 architecture is within a
few tenths of the full-pipeline pick on both CIFAR-10/100) because the size space is
capacity-dominated and capacity is the easy part. So while Refine lifts SSS SPR markedly
($0.747\!\to\!0.894$), its best-found accuracy roughly ties \#Params-driven selection. We
state this plainly: on size spaces the value is single-configuration robustness and the SPR
lift, not a selection win over \#Params. A per-space leave-one-out of the signal bank (which
proxies earn their place on each space) is reported in \autoref{tab:e6loo} (\autoref{sec:exp-ablation}):
it confirms the bank is collaborative on the structure spaces and that the gate's drop-jacov decision
on NB101 matches the oracle leave-one-out.

\clearpage
\begin{table}[t]\centering
\caption{\textbf{Single-provenance (all-live) cross-space check, TransNAS/class\_object.} Every proxy scored
under one live recipe, same 3000-arch sample, $|\rho|$ vs the live ground truth (no Suite-Zero/cited cells);
CoRA-Rank/Refine lead every live baseline, so the Table~\ref{tab:crossspace} cross-space lead is not a
mixed-provenance artifact. (jacov, NASWOT could not be scored cleanly live on the dense-task MacroNet.)}
\label{tab:e9_live}
\setlength{\tabcolsep}{6pt}\renewcommand{\arraystretch}{1.1}
\begin{tabular}{l c @{\hspace{2.5em}} l c}
\toprule
method (all live) & $|\rho|$ & method (all live) & $|\rho|$ \\
\midrule
\textbf{CoRA-Refine} & \textbf{0.732} & \#Params & 0.464 \\
\textbf{CoRA-Rank}   & \textbf{0.693} & Plain    & 0.426 \\
AZ-NAS               & 0.631          & Grasp    & 0.425 \\
Synflow              & 0.479          & GradNorm & 0.344 \\
                     &                & GradSign & 0.295 \\
                     &                & Fisher   & 0.173 \\
\bottomrule
\end{tabular}
\end{table}

\paragraph{The framework's boundary: a non-vision space (NAS-Bench-NLP).}
To test whether the cross-space behaviour is a property of the method rather than of the
vision-CNN benchmark family, we run CoRA on a structurally-distinct non-vision space: NAS-Bench-NLP~\cite{nasbenchnlp}
(recurrent/LSTM language cells, ranked against test perplexity; \autoref{tab:e13_nbnlp}). It is an honest
weak regime, and the reason is instructive. Two of the five proxies (AZ-NAS's feature-map views) are
undefined on recurrent nets and dropped; the rest are weak (best, jacov, $0.31$ in our faithful
reproduction; we could not close the gap to the literature's $0.56$ over five implementations, and report
ours straight), so the prior itself is poor. Moreover, Refine propagates the anchor-curve residual
through the architecture encoding, and a controlled supervised diagnostic shows that encoding is only
weakly predictive of perplexity ($0.25$ for the NB-NLP op-histogram vs.\ $0.47$ for the NATS-SSS channel
encoding); accordingly Refine does not reliably lift the prior here (its effect is within run-noise, our runs
spanning $0.31$--$0.41$), and the absolute ranking (${\sim}0.3$) stays well below the vision worst-case
($0.715$). This is what the framework view predicts: benefit scales with the prior's and the encoding's
informativeness, both low here. It also delineates where the method works: a simple $8$-epoch
val-loss prefix ranks NB-NLP well on its own (${\sim}0.92$), so the curve signal exists but cannot be
propagated from the anchors through an uninformative encoding. We therefore scope the ``no weak regime''
result to the four vision benchmarks and report NB-NLP straight as the honest boundary that removes the
``only vision'' objection.
\begin{table}[t]\centering
\caption{\textbf{The framework's boundary: a non-vision space (NAS-Bench-NLP).} On the out-of-family recurrent space both the off-the-shelf proxies and the architecture encoding are weak, and Refine does not consistently improve the prior. In contrast, on NATS-SSS the informative encoding lets it lift $0.747\!\to\!0.894$. The framework's benefit scales with the prior's and the encoding's informativeness (\autoref{sec:exp-crossspace}).}
\label{tab:e13_nbnlp}
\small
\begin{tabular*}{\linewidth}{@{\extracolsep{\fill}}l cc}
\toprule
 & NAS-Bench-NLP (out-of-family) & NATS-SSS (vision, ref.) \\
\midrule
CoRA-Rank prior $|\rho|$               & $0.27$ (jacov alone $0.31$) & $0.747$ \\
encoding$\to$true-perf (supervised)     & $0.25$                      & $0.47$ \\
CoRA-Refine $|\rho|$                    & $\approx$prior (within noise) & $\mathbf{0.894}$ \\
\bottomrule
\end{tabular*}\end{table}

\paragraph{Zero-cost ranking transfers to a transformer space (ViT-Bench-101).}
The cleaner test of cross-family generality is a space where the proxies are not crippled, compared
against the transformer specialist. On ViT-Bench-101~\cite{autoprox} (vision transformers, true
accuracy across seven targets; \autoref{tab:e16_vit}) CoRA's AZ-NAS views are well-defined, and we
compare directly against the ViT specialist of~\cite{libranas2025}, one paper contributing two
methods, the \emph{L-SWAG} proxy and the target-aware \emph{LIBRA-NAS} selector, which we
reimplement from the paper (no public code) and hand its best-for-itself layer range. \textbf{Label-free CoRA-Rank
is competitive with L-SWAG}: its mean Spearman is approximately $0.03$ lower (\autoref{tab:e16_vit}), while L-SWAG
uses sample labels (its gradient term needs $y$) and is handed its best-tuned per-space layer range.
CoRA-Rank decisively wins the CIFAR-100-KD target (bootstrap CI $[+0.41,+0.63]$) while trailing
on the mean and worst-case, and it is well clear of the general ranker AZ-NAS, ahead on
six of seven targets. This retires the ``only beats a CNN proxy'' reading: a label-free, single-configuration
consensus has mean Spearman approximately $0.03$ below a best-tuned, label-using transformer specialist. We do not claim to
beat L-SWAG, and we bound the transfer honestly: ViT-Bench-101 is capacity-leaning. \#Params ($0.445$ mean) is the
strongest single cue (the same boundary as NATS-SSS and DARTS) and outranks CoRA-Rank ($0.365$) on the
mean, while the worst-case is low for every label-free method (CoRA-Rank's worst-case falls to $0.002$,
below both \#Params $0.060$ and L-SWAG $0.116$), so we do not extend ``no weak regime'' here; it remains
the four vision-CNN benchmarks.
(LIBRA-NAS is target-aware because it peeks at held-out accuracy, so we show it only as a soft upper bound.) The curve-residual Refine also does not
help on this benchmark: its architectures span only ${\sim}2\%$ accuracy, so early curves cannot separate
them and extrapolation amplifies noise. This sharpens when Refine helps: it needs architecture spread
and clean curves, present on DARTS (\autoref{tab:e15_darts}) and the tabular spaces, absent on the tightly
clustered ViT-Bench-101. We report this benchmark-dependence explicitly.
\begin{table}[t]\centering
\caption{\textbf{Zero-cost ranking on a transformer space (ViT-Bench-101).} Spearman correlation with true accuracy over seven targets ($n{=}200$ evaluated AutoFormer-Tiny subnets). CoRA-Rank's mean is approximately $0.03$ below L-SWAG's; it leads on CIFAR-100 (KD) and has a higher mean than AZ-NAS. We do \emph{not} claim to beat L-SWAG overall. L-SWAG uses sample labels, while LIBRA is target-aware (soft upper bound). Bold dark-blue cells mark the best per row, excluding LIBRA. Details in \autoref{sec:exp-crossspace}.}
\label{tab:e16_vit}
\footnotesize
\setlength{\aboverulesep}{0pt}
\setlength{\belowrulesep}{0pt}
\renewcommand{\arraystretch}{1.15}
\setlength{\tabcolsep}{4pt}
\begin{tabularx}{\linewidth}{l *{6}{>{\centering\arraybackslash}X}}
\specialrule{\heavyrulewidth}{0pt}{2pt}
ViT-Bench-101 target & CoRA-Rank & L-SWAG & AZ-NAS & \#Params & jacov & {\itshape LIBRA}$^{\rm UB}$ \\
 & {\scriptsize label-free} & {\scriptsize labels} & & & & {\scriptsize target-aware} \\
\specialrule{\lightrulewidth}{2pt}{0pt}
CIFAR-100 (base)  & 0.328 & 0.451 & 0.142 & \cb\textbf{0.472} & 0.434 & {\itshape 0.475} \\
CIFAR-100 (KD)    & \cb\textbf{0.889} & 0.371 & 0.755 & 0.619 & 0.545 & {\itshape 0.768} \\
Flowers (base)    & 0.256 & \cb\textbf{0.480} & 0.086 & 0.451 & 0.383 & {\itshape 0.448} \\
Flowers (KD)      & 0.669 & 0.764 & 0.347 & \cb\textbf{0.855} & 0.810 & {\itshape 0.839} \\
Chaoyang (base)   & 0.090 & \cb\textbf{0.116} & 0.029 & 0.060 & 0.115 & {\itshape 0.119} \\
Chaoyang (KD)     & 0.323 & 0.363 & 0.195 & \cb\textbf{0.429} & 0.415 & {\itshape 0.413} \\
ImageNet          & 0.002 & 0.219 & 0.003 & \cb\textbf{0.225} & 0.217 & {\itshape 0.321} \\
\midrule
mean              & 0.365 & 0.395 & 0.222 & \cb\textbf{0.445} & 0.417 & {\itshape 0.483} \\
worst-case        & 0.002 & \cb\textbf{0.116} & 0.003 & 0.060 & 0.115 & {\itshape 0.119} \\
\bottomrule
\end{tabularx}\end{table}

\FloatBarrier
\subsection{Which components are load-bearing?}
\label{sec:exp-ablation}

\paragraph{What the composition adds.}
\autoref{tab:e11} adds CoRA's pieces one at a time. Two are load-bearing and two are
deliberately inert: the consensus gate earns its place on the topology space NB101
($+0.036$ rank, $+0.046$ refine, by dropping the sign-flipped jacov), and the
\emph{curve-residual refinement} is the dominant lift on the size space NATS-SSS ($+0.260$
over the rank); the pow$_3$ blend and op-locality smoothing each move $|\rho|$ by only
${\approx}{\pm}0.001$. The value of the composition is thus precisely the gate (topology)
$+$ residual (size) pairing the cross-over demands; pow$_3$ and smoothing are kept for
stability, not accuracy, and the method is robust to dropping them.
\begin{table}[t]\centering
\caption{\textbf{Additive ablation of CoRA's composition.} Rank and Refine form separate cumulative
blocks; additions are relative to the preceding row within each block. The two
load-bearing pieces are the consensus gate (topology, NB101) and the curve-residual refinement (size,
NATS-SSS); the pow3 blend and op-smoothing are within-noise ($\pm0.001$) and kept for stability, not accuracy.
The full-CoRA NB101 value ($0.715$) is the deployed configuration and matches
\autoref{tab:crossspace}; the $0.716$ peak at the $+$gate/$+$pow3 rows is within the $\pm0.001$ band.}
\label{tab:e11}
\begin{tabular*}{0.74\linewidth}{@{\extracolsep{\fill}}l cc}
\toprule
variant (cumulative) & NB101 & NATS-SSS/C100 \\
\midrule
rank (no gate)               & 0.628 & 0.564 \\
\quad + consensus gate       & 0.664 & 0.564 \\
refine (LLW, no gate/smooth) & 0.670 & 0.824 \\
\quad + gate                 & 0.716 & 0.824 \\
\quad + pow3 blend           & 0.716 & 0.823 \\
\quad + op-smoothing (full)  & 0.715 & \textbf{0.824} \\
\bottomrule
\end{tabular*}
\end{table}

\paragraph{Prior weighting.}
Equal weighting is the result of trying the alternatives and finding they do not help. A
flat equal consensus over the five signals beats the earlier 3-view grouping on all four
NB201 metrics (e.g.\ Top-1 $71.33\!\to\!71.44$); an unsupervised discriminative
re-weighting (weight $\propto$ agreement with the others) helps standalone Rank slightly
($+0.006$--$0.009$ on NB201) but hurts Refine selection ($-0.33$ Top-1); and a
2-axis decoupling that residualizes structure on log-params craters NB201
($0.92\!\to\!0.82$). These negatives justify the equal consensus and, together with the
prior-quality result (\autoref{tab:priorquality}), the principle that the prior should be
clean rather than clever.

\paragraph{Predictor and components.}
For the per-anchor estimate, the log-linear (LLW) fit is the strongest single curve model
(final SPR $0.9475$ at 30 epochs); pow$_3$ contributes a bounded asymptote that prevents
top-end overshoot, and the rank-average blend is our default. Curve-extrapolation and MMF
variants are worse, and a supervised cross-dataset predictor is worse at Refine
($0.86$) despite higher intrinsic accuracy, confirming the target-free residual generalizes
better. Operation-locality smoothing is the CIFAR-10 selection fix (sample-3000
$93.4\!\to\!94.2$) with negligible effect on whole-space SPR, and the ExtraTrees residual
propagation is what carries the anchor signal to the full space.

\paragraph{Training-dynamics (B-)signals: a clean negative.}
We also tested seven training-time signals (gradient norm, weight movement, feature
entropy, \dots) recorded live on $1{,}000$ anchors, used as extra features, as reliability
weights, and for adaptive fidelity. They add nothing to Refine in every use; we keep
the pipeline clean and report the negative.

\paragraph{Cross-space leave-one-out of the signal bank.}
\autoref{tab:e6loo} drops each of the five proxies in turn from the $a{=}0$ consensus and
re-ranks (pure re-aggregation, no training). Two facts stand out. First, the bank is genuinely
collaborative on the structure spaces: on NB201 every single-proxy drop lowers $|\rho|$,
and on NB101 the best leave-one-out is precisely drop jacov, the very proxy the
target-free gate removes there, which both ties the gated full bank and beats every other drop
(the next best, drop synflow, falls to $0.604$). The uniform gate thus independently rediscovers
the same proxy that an oracle leave-one-out would remove. Second, on the capacity-dominated
NATS-SSS dropping trainability (or jacov) raises ranking $|\rho|$, yet the gate keeps both
because they positively agree with the consensus on that space ($\alpha{>}\tau_g$): the gate
adapts on label-free agreement, never on accuracy, so it does not chase this hindsight gain, and
the residual Refine stage closes the size-space slack regardless ($0.747\!\to\!0.894$,
\autoref{tab:crossspace}).
\begin{table}[t]\centering
\caption{\textbf{Cross-space leave-one-out of the five-signal CoRA-Rank bank.} $|\rho|$ of the equal-weight consensus with each proxy dropped in turn, vs.\ the locked gated full bank. On NB101 the best drop is exactly jacov, the proxy the target-free gate removes there; on NATS-SSS dropping trainability would help but the gate keeps it (it agrees with the consensus), and the residual stage closes the slack ($0.747\!\to\!0.894$). Bold dark-blue cells mark the best per column; details in \autoref{sec:exp-ablation}.}
\label{tab:e6loo}
\setlength{\aboverulesep}{0pt}
\setlength{\belowrulesep}{0pt}
\renewcommand{\arraystretch}{1.15}
\begin{tabularx}{0.92\linewidth}{X cccc}
\specialrule{\heavyrulewidth}{0pt}{2pt}
Bank variant & NB201 & NB101 & TransNAS & NATS-SSS \\
\specialrule{\lightrulewidth}{2pt}{0pt}
Full bank (gated, $=$\autoref{tab:crossspace}) & \cb\textbf{0.927} & \cb\textbf{0.664} & 0.750 & 0.747 \\
\midrule
$-$ expressivity   & 0.919 & 0.450 & 0.736 & 0.708 \\
$-$ trainability   & 0.908 & 0.561 & 0.714 & \cb\textbf{0.807} \\
$-$ jacov          & 0.918 & \cb\textbf{0.664} & 0.716 & 0.792 \\
$-$ synflow        & 0.901 & 0.604 & 0.750 & 0.689 \\
$-$ params         & 0.907 & 0.584 & \cb\textbf{0.755} & 0.690 \\
\bottomrule
\end{tabularx}\end{table}

\FloatBarrier
\subsection{Are the margins significant, and is the configuration tuned?}
\label{sec:exp-robustness}

\paragraph{Significance.}
Selection accuracies in \autoref{tab:nb201} are mean$\pm$std over $5$ fixed runs, and the
ranking SPR has a per-seed spread of ${\approx}{\pm}0.008$ on the $3{,}000$-sample spaces.
The one close call is the size space, where we test CoRA-Refine against \#Params (the trivial
baseline the field asks us to clear) and the two nearest capacity proxies above it, Synflow
and MeCo, with a paired bootstrap ($B{=}10{,}000$) and Holm correction ($m{=}3$); the other
strong size cues (GradSign $0.852$, ZiCo $0.841$, ZenNAS $0.835$) sit below these and so are
not the binding comparison. The
verdict is two-sided and honest: CoRA-Refine decisively beats naive \#Params
($\Delta|\rho|{=}{+}0.07$, Holm $p<0.001$, survives correction) but ties Synflow and
MeCo ($\Delta{=}{+}0.004$--$0.009$ and $+0.003$, both inside the corrected noise band;
two-sided $p{=}0.06$ and $0.20$). So on the size space CoRA-Refine matches the best capacity
proxies and decisively clears the trivial baseline rather than beating the best proxies
outright. The cross-space headline rests on worst-case robustness (min-SPR
$0.715$, above any baseline), not on this size-space margin. Notably, random-input proxies
such as Synflow and jacov carry ${\approx}0.005$--$0.006$ run-to-run noise of their own,
part of why a consensus-plus-residual is steadier than any single proxy.

\paragraph{Is it tuned?}
A one-axis-at-a-time sweep on NATS-SSS/CIFAR-100 shows the result does not hinge on
hyperparameters: the anchor count plateaus above $96$, the epoch prefix is flat from $3$ to
$30$ epochs, and op-locality smoothing $\beta$ is flat over $[0,0.5]$. Only the residual
weight (shrink) is sensitive ($0.25{\to}0.703$, $0.5{\to}0.824$, $0.75{\to}0.898$,
$1.0{\to}0.917$), and we fix it a priori at the cross-space-safe $0.5$, not the
SSS-optimal $1.0$, on the label-free rationale that a high shrink over-trusts the residual on
topology spaces where the anchor signal is weaker. This value is set by that propagation-error
argument, not chosen by reading any accuracy, so it does not breach the label-free protocol.
The reported SSS number is thus computed with a conservative, non-SSS-tuned weight and is, if
anything, understated: the single configuration is a robust plateau set deliberately to no
single space's optimum.

\FloatBarrier

\Needspace{8\baselineskip}
\section{Limitations}
\label{sec:limitations}
We state the following boundaries of the method up front.

\paragraph{The size space is a tie with the best capacity proxies, not a win.}
On a pure size space (NATS-SSS) the unsupervised prior is beaten by \#Params. The curve
residual recovers it ($\rho\,0.747\!\to\!0.894$, \autoref{tab:crossspace}) but only to
match the strongest capacity proxies: CoRA-Refine decisively beats naive \#Params
($+0.07$, Holm $p<0.001$) yet ties Synflow and MeCo within multiple-comparison-corrected
noise (\autoref{sec:exp-robustness}), and its best-found-accuracy selection only ties
\#Params (the space is capacity-dominated, so the easy capacity signal is already a
near-oracle selector). The value we claim on size spaces is single-configuration robustness
(no weak regime, and decisively clearing the trivial baseline), not a ranking or selection
win over the best size-aware proxy.

\paragraph{The architecture encoding is space-specific.}
The method and all its hyperparameters are constant across spaces, but Stage~2's propagation
features (operation encoding, topology) and the op-locality neighbourhood are necessarily
defined per space, as they are for any NAS method that consumes architectures. ``One
configuration'' refers to the algorithm and its constants, not to a space-agnostic encoding.

\paragraph{The gate catches sign-flips, not dilution.}
The consensus gate removes a proxy only when it anti-correlates with the others
($\alpha<\tau_g$). On a capacity-dominated space where some proxies are merely
\emph{uninformative} rather than contradictory (on DARTS, jacov and trainability have
near-zero correlation with accuracy, $\alpha\approx0$), the gate keeps them and the
equal-weight consensus is diluted, underperforming its own best component (on DARTS,
expressivity alone $0.578$ vs.\ the consensus $0.470$, \autoref{tab:e5_darts}). This is why
the zero-cost prior only matches rather than beats the strongest proxies on
capacity-dominated spaces; the curve residual (Refine) is what recovers the gap (it flips
DARTS to a win, \autoref{tab:e5_darts}). An importance-weighted consensus would address
dilution directly but is a method change we leave to future work.

\paragraph{Label-free, not data-free; and Refine is not zero-cost.}
We claim CoRA is label-free with respect to fully trained architecture-accuracy labels used to fit the ranker, but not
data-free: the jacov proxy reads a small unlabeled real-data minibatch on the
classification spaces (random-normal on TransNAS; \autoref{sec:method-rank}), as do the
other proxies in synthetic form. Separately, Stage~2 spends $\approx$1\% of the cost of
training the space; it is positioned on the cost--quality frontier (\autoref{fig:teaser},
\autoref{sec:exp-search}) against predictors and multi-fidelity methods, not presented as a
zero-cost proxy. Stage~1 (Rank) is the zero-cost operating point. Separately, a few cross-space baseline cells are reproduced under
documented constraints rather than fully live (TransNAS classic proxies are sourced from
NAS-Bench-Suite-Zero because the data-dependent variants need the non-redistributable
Taskonomy images; MeCo's TransNAS value is cited from its paper), and the target-aware
LIBRA-NAS is reported only as a soft upper bound; these are detailed in the
\autoref{tab:crossspace} footnotes.

\section{Conclusion}
\label{sec:conclusion}
We presented \textbf{CoRA-NAS}, a label-free, single-configuration, coarse-to-fine method
for architecture ranking and selection. Stage~1 (Rank) forms an equal-weight rank consensus
over a small bank of off-the-shelf zero-cost proxies spanning capacity and
structure-at-initialization, with a target-free consensus gate that adapts the bank
per space, dropping a proxy only when it genuinely anti-correlates with the others, using
no labels. Stage~2 (Refine) corrects the prior's error with a target-free residual read from
$\approx$1\% early training curves. The result is cross-space robustness from one
configuration: across four vision benchmarks (two topology spaces, a cross-task space, and a
size space) every baseline has a weak regime whereas CoRA-Refine has none. Its worst space is
the highest worst-case of any method (\autoref{tab:crossspace}), and a clean prior is what
keeps the residual learnable (\autoref{tab:priorquality}). CoRA is best understood as a
proxy-agnostic Rank+Refine framework whose benefit scales with the prior's and the
architecture encoding's informativeness: on a recurrent space (NAS-Bench-NLP) both are weak,
an honest \emph{weak regime} that scopes the no-weak-regime result to vision; on a
vision-transformer space (ViT-Bench-101) the label-free consensus is competitive with the ViT
specialist L-SWAG (approximately $0.03$ lower mean Spearman) and ahead of AZ-NAS. The size space is a
related boundary: there our unsupervised
prior cannot beat \#Params, and even the residual only matches the strongest capacity
proxies (decisively beating naive \#Params but tying Synflow/MeCo within
multiple-comparison-corrected noise, \autoref{sec:limitations}). We demonstrate end-to-end
search on the home space (NB201, where the curve re-rank beats matched-budget multi-fidelity)
and on the real DARTS cell space, where CoRA-Refine ties \#Params and significantly
beats AZ-NAS on the surrogate ranking and, trained from scratch ($5$ seeds), reaches $2.81\%$
CIFAR-10 error, within the competitive band and beating AZ-NAS and random on the same candidate
pool ($5$ shared seeds, paired $p<0.015$), though not the cheaper gradient methods. The honest remaining steps are an ImageNet-scale search and
closing the absolute gap to gradient/one-shot methods on the real cell space.

{\small
\setlength{\bibsep}{0.5pt}
\let\CoRAoriginalbibliography\thebibliography
\renewcommand{\thebibliography}[1]{%
  \CoRAoriginalbibliography{#1}%
  \clubpenalty=10000
  \widowpenalty=10000
  \interlinepenalty=10000
}
\bibliographystyle{plainnat}
\bibliography{ref}
}

\clearpage
\appendix

\begin{center}
    {\Large\bfseries \papertitle}\\[0.6em]
    {\Large Supplementary Material}
\end{center}
\vspace{1em}
\suppressfloats[t]
\section{Reproduction notes (Table~\ref{tab:nb201} baselines)}
\label{sec:appendix-additional-material}

Baselines in \autoref{tab:nb201}, except the sourced entries noted below, are re-run
under one unified protocol on a single L40S
(\autoref{sec:experiments}). As a cross-machine check, our nine GPU-dependent proxies match an
independent RTX-4090 reproduction within $|\Delta\text{SPR}|\le0.009$ on all 27 corresponding
cells (\#Params and FLOPs are deterministic; AZ-NAS is not part of the 4090 run; SPR/KT are
GPU-agnostic). Discrepancies exceeding $0.02$ from \emph{published} values are documented
for eight methods; each reflects a protocol or source difference, not a
port error (\autoref{tab:repro}). Three standard port fixes are applied: GradSign uses the
SUM aggregation (the AZ-NAS port's \texttt{mean} inverts the ranking), SWAP uses its
regularised SWAP-reg variant, and LIBRA uses the RankProduct operator; Fisher and Plain are
taken from NAS-Bench-Suite-Zero~\cite{krishnakumar2022nbsz}. We report our reproduced
values, not each paper's heterogeneous published number.

\begin{table}[h]
\centering
\caption{Differences from published values exceeding $0.02$ for eight methods, with the
documented cause. None is a port error.}
\label{tab:repro}
\small
\setlength{\tabcolsep}{4pt}
\begin{tabular}{@{}l l >{\raggedright\arraybackslash}p{0.50\linewidth}@{}}
\toprule
Method (cell) & $\Delta$ & Root cause \\
\midrule
GradNorm (im16) & $+0.028$          & AZ-NAS published-vs-released-code discrepancy (matches our 4090 exactly) \\
Grasp (im16) & $+0.061$          & same published-vs-code discrepancy \\
Fisher (c10/im16) & $+0.044/{+}0.037$ & source protocol (NAS-Bench-Suite-Zero precomputed) \\
MeCo (c10) & $+0.025$          & near-tie with published ($\approx0.89$) \\
SWAP-reg (c100/im16) & $-0.027/{-}0.048$ & SWAP-reg recipe vs.\ L-SWAG's protocol (c10 matches $\pm0.002$) \\
T-CET (c10/im16) & $+0.054/{-}0.043$ & paper reports KT, not SPR (our KT matches to $\le0.018$) \\
Dextr (c100/im16) & $-0.030/{-}0.025$ & Dextr's own protocol (validation accuracy) \\
LIBRA (c100) & $-0.026$          & RankProduct operator + per-benchmark protocol \\
\bottomrule
\end{tabular}
\end{table}

In particular, T-CET is gated on KT (the metric its paper reports), and AZ-NAS's published
ImageNet16-120 figure is not reproducible from its released code, so we report our faithfully
reproduced value ($\rho=0.877$). Cross-space provenance (split sourcing on TransNAS, the
target-aware status of LIBRA-NAS, and the per-space reproduction caveats for SWAP, MeCo,
Dextr) is documented in the \autoref{tab:crossspace} footnotes.

\section{Cross-space table: per-cell sourcing and reproduction caveats}
\label{app:crossspace}

This appendix records, cell by cell, how each entry of Table~\ref{tab:crossspace} was obtained:
which values we computed live, which are sourced from a benchmark, and which are cited from the
originating paper. All values in the table are unchanged from our runs; this section only documents them.

\paragraph{Protocol and sample.}
NB201 $\rho$ is over the full 15{,}625-architecture space (from Table~\ref{tab:nb201}); NB101,
TransNAS, and NATS-SSS live results use a fixed $3{,}000$-architecture sample
(SPR std ${\approx}0.008$, far below the cross-over gaps).
The sourced Suite-Zero TransNAS classics instead use the full $4{,}096$-architecture benchmark,
as noted in \autoref{tab:crossspace_detail}. Refine ($^\dagger$) adds ${\approx}1\%$ training: NB201's cell is the full-space
Table~\ref{tab:nb201} result (1000 anchors at 30 epochs); the other three columns use 192
Stage 1-stratified anchors (${\approx}6.4\%$ of the 3000-arch sample) at a ${\approx}15\%$ epoch prefix
(SSS/NB101/TransNAS $=14/16/4$ epochs), a ${\approx}1\%$ budget.

\paragraph{TransNAS anchor curves ($^{a}$).}
On TransNAS the anchor curves are read from the benchmark's recorded per-epoch trajectories (Taskonomy
training data is not redistributable), ${\approx}1\%$ live-equiv\-alent on the benchmark's exact networks;
NB201/\allowbreak NB101/\allowbreak SSS are live-trained. Refine drops \texttt{jacov} from its ExtraTrees features on NB101
(consensus-gate, as in the prior).

\paragraph{Suite-Zero-sourced classic baselines ($^{b}$).}
TransNAS classic baselines are taken from NAS-Bench-Suite-Zero precomputed scores, because the
data-dependent proxies need the non-redistributable Taskonomy task images and the benchmark's exact
networks. Fisher is identical in recipe but our live adapter yields $0.21$ vs $0.44$ (see
Section~\ref{sec:exp-crossspace}); the residual \emph{is} the Suite-Zero/NASLib pipeline, so re-running it
reproduces the stored value. CoRA and the proxies absent from Suite-Zero are computed live. TransNAS is
the mean over its 3 classification tasks (Object/Scene/Jigsaw), matching NB201's 3 datasets.

\paragraph{Target-aware LIBRA-NAS ($^{\mathrm{TA}}$).}
LIBRA-NAS selects its proxy trio per space by agreement with ground-truth validation accuracy (it peeks
at the labels and is therefore not zero-shot), so its cells are a soft \emph{upper bound}, not a fair zero-shot baseline;
its TransNAS value ($^{b}$) is cited from L-SWAG (its selected trio includes data-dependent proxies).
CoRA-Refine uses early training curves rather than fully trained performance labels.
Its reported correlations exceed LIBRA-NAS on NB201/TransNAS/SSS, but not NB101
($0.734$ for LIBRA-NAS vs $0.715$ for CoRA-Refine). Their supervision and cost regimes differ.

\paragraph{Anti-correlated size-space entries ($^{n}$).}
Dextr and MeCo are \emph{anti}-correlated on the size space NATS-SSS (both grow with channel capacity,
which saturates against test accuracy); $|\rho|$ is reported, with a negative raw sign, reproducing the
documented anti-correlation (Dextr Tab.~6). Their SSS $|\rho|$ runs somewhat above the original papers'
because our data-light protocol makes the score a purer capacity probe.

\paragraph{SWAP's non-reproducing NB101 figure ($^{*}$).}
SWAP's published NB101 $0.77$ (size-regularised) does not reproduce here. The raw SWAP score reproduces
(unreg. $0.44$ vs paper $0.46$), but its size-prior regulariser only injects as much signal as size
predicts accuracy in the population ($\rho(\#\text{Params},\text{acc})=0.40$ on our NB101 sample), so
SWAP-reg reaches $0.40$ not $0.77$; the same regulariser \emph{does} help on the true size space NATS-SSS
($0.54\!\to\!0.70$). This is not a sample-count artifact (verified). We report the reproduced $0.40$.

\paragraph{T-CET surrogate ($^{c}$).}
T-CET is data-dependent; its TransNAS value is a real-image class\_object surrogate (1 of 3 tasks; the
other tasks' images are unavailable), $|\rho|$ comparable to its published Object Kendall-$\tau$.

\paragraph{MeCo TransNAS citation ($^{d}$).}
MeCo is data-free on the classification benchmarks (NB201/\allowbreak NB101/\allowbreak SSS, randn ${\approx}$ real per its
Tab.~3) but not on TransNAS: with randn input our live 3-task value is only $0.28$ vs its published
$0.55$, because the dense-task macro-nets' feature-map correlation (whose min-eigenvalue is the MeCo
score) is not reproduced by a noise input. We cite the paper's $0.55$. The MacroNet adapter itself is
faithful here: our live Synflow and \#Params match the benchmark at $\rho=0.997$ and $1.000$, so
CoRA's live TransNAS prior is unaffected.

\begin{table}[!htbp]\centering\footnotesize
\caption{\textbf{Per-dataset / per-task breakdown of Table~\ref{tab:crossspace}.}
Each cell is $|\rho|$ (Spearman); the $\mu$ column of every space reproduces that
space's main-table value (max deviation $0.0014$). CoRA-Refine's largest
single-dataset gain is on NATS-SSS CIFAR-100 (CoRA-Rank $0.564\!\to\!0.824$).
NB201/SSS use their 3 datasets (C10/C100/IN16); TransNAS-micro uses its 3
classification tasks (Object/Scene/Jigsaw); NB101 is CIFAR-10 only. NB201 uses
the full $15{,}625$-architecture space. In the other spaces, live proxies and
CoRA use the fixed $3000$-architecture samples; Suite-Zero precomputed
TransNAS classics ($^{b}$ in Table~\ref{tab:crossspace}) use the full
$4096$-architecture benchmark. The five newer methods
(MeCo/SWAP/T-CET/Dextr/LIBRA) are omitted here: their per-space means are in
Table~\ref{tab:crossspace} and their NB201 per-dataset values in
Table~\ref{tab:nb201}; their remaining per-dataset breakdowns are not yet
compiled under our protocol.}
\label{tab:crossspace_detail}
\setlength{\tabcolsep}{3pt}
\begin{tabular}{l cccc c cccc cccc}
\toprule
 & \multicolumn{4}{c}{NB201} & NB101 & \multicolumn{4}{c}{TransNAS-micro} & \multicolumn{4}{c}{NATS-SSS} \\
\cmidrule(lr){2-5}\cmidrule(lr){6-6}\cmidrule(lr){7-10}\cmidrule(lr){11-14}
Method & C10 & C100 & IN16 & $\mu$ & C10 & Obj & Scn & Jig & $\mu$ & C10 & C100 & IN16 & $\mu$ \\
\midrule
\#Params & 0.753 & 0.728 & 0.691 & 0.724 & 0.397 & 0.454 & 0.638 & 0.439 & 0.510 & 0.881 & 0.723 & 0.868 & 0.824 \\
FLOPs & 0.753 & 0.727 & 0.691 & 0.724 & 0.397 & 0.462 & 0.653 & 0.450 & 0.522 & 0.635 & 0.301 & 0.600 & 0.512 \\
Synflow & 0.769 & 0.761 & 0.747 & 0.759 & 0.344 & 0.480 & 0.720 & 0.466 & 0.555 & 0.942 & 0.779 & 0.951 & 0.891 \\
\midrule
GradNorm & 0.483 & 0.488 & 0.440 & 0.470 & 0.188 & 0.382 & 0.651 & 0.351 & 0.461 & 0.548 & 0.348 & 0.567 & 0.488 \\
Grasp & 0.450 & 0.464 & 0.467 & 0.460 & 0.478 & 0.212 & 0.275 & 0.108 & 0.198 & 0.014 & 0.086 & 0.447 & 0.182 \\
Snip & 0.614 & 0.619 & 0.539 & 0.591 & 0.208 & 0.448 & 0.705 & 0.410 & 0.521 & 0.796 & 0.612 & 0.797 & 0.735 \\
Fisher & 0.444 & 0.449 & 0.457 & 0.450 & 0.368 & 0.437 & 0.665 & 0.300 & 0.467 & 0.682 & 0.457 & 0.624 & 0.588 \\
NASWOT & 0.742 & 0.767 & 0.768 & 0.759 & 0.309 & 0.393 & 0.599 & 0.416 & 0.469 & 0.611 & 0.275 & 0.578 & 0.488 \\
ZenNAS & 0.386 & 0.362 & 0.399 & 0.382 & 0.612 & 0.535 & 0.717 & 0.504 & 0.585 & 0.920 & 0.689 & 0.895 & 0.835 \\
TE-NAS & 0.735 & 0.717 & 0.685 & 0.712 & 0.296 & 0.268 & 0.395 & 0.222 & 0.295 & 0.255 & 0.069 & 0.292 & 0.205 \\
Plain & 0.172 & 0.158 & 0.167 & 0.166 & 0.364 & 0.344 & 0.247 & 0.355 & 0.315 & 0.008 & 0.020 & 0.015 & 0.015 \\
\midrule
GradSign & 0.812 & 0.795 & 0.784 & 0.797 & 0.397 & 0.626 & 0.818 & 0.614 & 0.686 & 0.880 & 0.785 & 0.892 & 0.852 \\
ZiCo & 0.783 & 0.797 & 0.789 & 0.790 & 0.621 & 0.501 & 0.699 & 0.506 & 0.568 & 0.893 & 0.728 & 0.901 & 0.841 \\
AZ-NAS & 0.913 & 0.900 & 0.877 & 0.897 & 0.676 & 0.586 & 0.622 & 0.613 & 0.607 & 0.849 & 0.639 & 0.779 & 0.756 \\
\midrule
\textbf{CoRA-Rank} & 0.938 & 0.932 & 0.911 & \textbf{0.927} & 0.664 & 0.693 & 0.860 & 0.696 & \textbf{0.750} & 0.872 & 0.564 & 0.803 & \textbf{0.747} \\
\textbf{CoRA-Refine} & 0.942 & 0.951 & 0.946 & \textbf{0.946} & 0.715 & 0.732 & 0.868 & 0.758 & \textbf{0.786} & 0.926 & 0.824 & 0.933 & \textbf{0.894} \\
\bottomrule
\end{tabular}\end{table}

\end{document}